\documentclass{article}
\let\irsimaddcontentsline\addcontentsline
\usepackage{iclr2027_conference,times}
\let\addcontentsline\irsimaddcontentsline

\usepackage{amsmath}
\usepackage{amssymb}
\usepackage{amsfonts}
\usepackage{booktabs}
\usepackage{multirow}
\usepackage{graphicx}
\usepackage{xcolor}
\definecolor{citationSkyBlue}{RGB}{86,180,233}
\usepackage{colortbl}
  \usepackage[
    colorlinks=true,
    citecolor=citationSkyBlue,
    linkcolor=citationSkyBlue,
    urlcolor=blue
  ]{hyperref}
\usepackage{url}

\definecolor{tablebest}{rgb}{0.78,0.70,0.94}
\definecolor{tablesecond}{rgb}{0.90,0.87,0.97}

\newcounter{subfigurepanel}[figure]

\newcommand{\figpanellabel}[1]{\refstepcounter{subfigurepanel}\label{#1}}

\title{IR-SIM: A Lightweight Declarative Simulator for Navigation Learning and Benchmarking}
\author{
  \textbf{Ruihua Han}$^{1}$ \quad
  \textbf{Shuai Wang}$^{2}$ \quad 
  \textbf{Chengyang Li}$^{1}$ \quad 
  \textbf{Rui Gao}$^{3}$ \quad \\
  \textbf{Xinyi Wang}$^{4}$ \quad  
  \textbf{Zhe Liu}$^{1}$ \quad 
  \textbf{Guoliang Li}$^{5}$ \quad
  \textbf{Yupu Lu}$^{1}$ \quad \\
    \textbf{Qi Hao}$^{3}$ \quad
    \textbf{Jia Pan}$^{1}$ \quad
    \textbf{Hengshuang Zhao}$^{1}$ \quad \\
  \normalfont\textsuperscript{1} The University of Hong Kong \\ 
  \textsuperscript{2} Shenzhen Institutes of Advanced Technology \\ 
  \textsuperscript{3} Southern University of Science and Technology \\
  \textsuperscript{4} University of Michigan \quad
    \textsuperscript{5} University of Macau \\
}
\iclrfinalcopy
\hypersetup{
  pdftitle={IR-SIM: A Lightweight Declarative Simulator for Navigation Learning and Benchmarking},
  pdfauthor={Ruihua Han, Shuai Wang, Chengyang Li, Rui Gao, Xinyi Wang, Zhe Liu, Guoliang Li, Yupu Lu, Qi Hao, Jia Pan, Hengshuang Zhao},
  pdfsubject={Preprint},
  pdfkeywords={robotics, navigation, simulation, reinforcement learning, benchmarking}
}

\begin{document}
\maketitle
\lhead{Preprint}

\begin{abstract}
Developing navigation policies requires simulation scenarios that support repeatable training and evaluation. Despite the availability of numerous simulators, constructing diverse navigation scenarios often involves writing simulator-specific code, using complex graphical interfaces, performing substantial manual configuration, or relying on resource-intensive computing platforms, making scenarios difficult to reproduce and reuse across experiments.
To this end, we develop the Intelligent Robot Simulator (IR-SIM), a lightweight declarative simulator implemented as a Python library to support navigation learning and benchmarking through rapid construction of reusable and diverse scenarios and efficient execution on accessible platforms. IR-SIM represents scenarios as human-readable YAML configurations, in which necessary objects, behaviors, sensors, maps, and environment parameters can be flexibly specified and composed.
This declarative representation makes IR-SIM friendly to large language model (LLM)-powered agents, enabling them to compose and modify executable navigation scenario files through the provided agent skills, instead of writing a large amount of code that is difficult to reuse.
Despite its lightweight implementation, IR-SIM provides the essential components for navigation simulation, while retaining interfaces to high-fidelity simulators for downstream validation.
Experiments demonstrate that IR-SIM runs up to several hundred times faster on the evaluated CPU platform, while agent skills reduce mean LLM-based scenario construction time by more than $50\%$ across both evaluated models. The experiments further show ability to support reproducible benchmarking and RL-based navigation policy learning.
\end{abstract}

\section{Introduction}
Navigation scenario construction is a necessary step in developing and evaluating navigation algorithms in robotics research. It is increasingly important with the continued development of reinforcement learning (RL)-based navigation approaches, which typically require large numbers of diverse and reusable environments for policy training~\citep{fan2020distributed,xu2025navrl}.
Numerous general-purpose high-fidelity simulators, such as Gazebo~\citep{koenig2004design}, CARLA~\citep{dosovitskiy_carla_2017}, and Isaac Sim~\citep{NVIDIA_Isaac_Sim}, provide realistic environments with 3D physics-based engines to support various navigation tasks~\citep{li2024edge, han2025neupan, xu2026navrl++}.
However, constructing scenarios can require learning simulator-specific APIs, configuring scene assets, and implementing agent behaviors. The computational demands of high-fidelity simulation can also limit repeated training and evaluation on accessible hardware. These challenges motivate simulators that combine rapid scenario construction with efficient execution.

\label{sec:introduction}
\begin{figure}[t]
    \centering
    \includegraphics[width=0.95\linewidth]{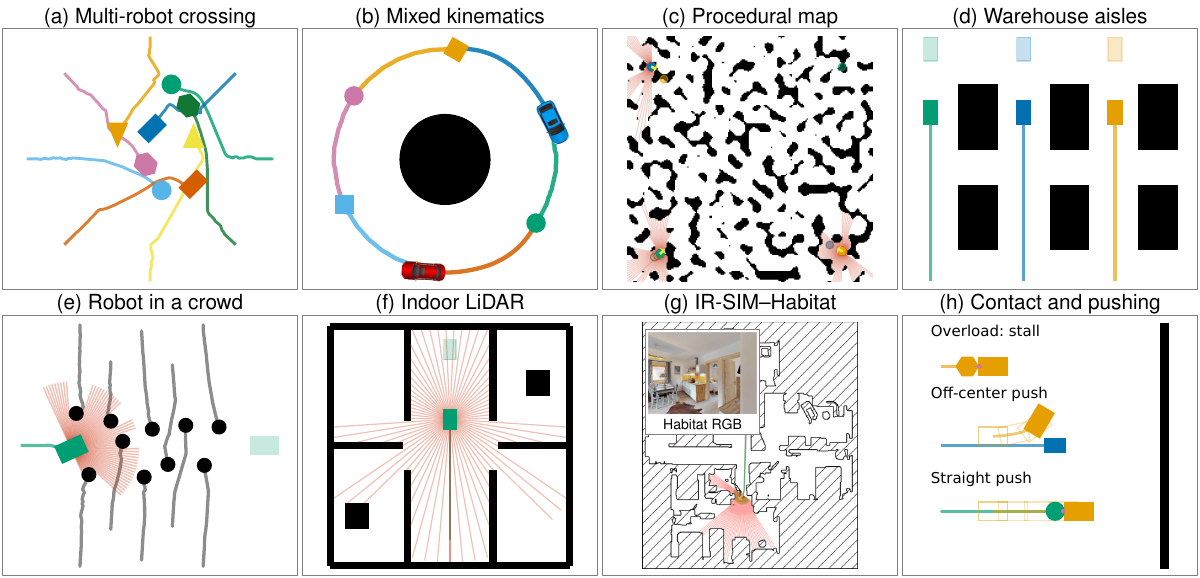}
    \caption{Navigation scenarios generated in IR-SIM. Declarative YAML configurations compose robot kinematics, geometry, agent behaviors, and sensing to support multi-agent encounters, social navigation, structured and procedural environments, and contact interactions.}
    \label{fig:scenario_gallery}
\end{figure}

Thus, to reduce scenario construction complexity and improve simulation efficiency, lightweight simulators have been developed for specific navigation tasks, such as multi-agent reinforcement learning~\citep{bettini2022vmas}, social navigation~\citep{chen2019crowdrobot}, multi-vehicle simulation~\citep{blanco2023mvsim}, differential-drive robot navigation~\citep{rutherford2024noregrets}, and multi-agent driving~\citep{kazemkhani2025gpudrive}.
These simulators provide efficient solutions for their target applications, but their designs typically focus on particular navigation subdomains or a limited set of robot models and geometries, or depend on specific software frameworks (e.g., ROS, PyTorch) or hardware prerequisites (e.g., GPUs). These constraints make it difficult to generate diverse scenarios and reuse them across different navigation learning and evaluation workflows.

Meanwhile, large language models (LLMs) have been used to generate robotic simulation tasks and training supervision~\citep{wang_gensim_2024,wang_robogen_2024}.
LLM agents can construct scenes through simulator APIs, but successful authoring requires both correct API use and the intended runtime behavior. Generating simulator-specific scene and controller code can increase the amount of code that must be produced, inspected, and maintained. A compact scenario representation backed by reusable behavior implementations offers a way to simplify this authoring process for both users and LLM agents.

To address these challenges, we develop the \textbf{Intelligent Robot Simulator (IR-SIM)}, a lightweight declarative simulator implemented as a Python library for navigation learning and benchmarking. IR-SIM is fast, extensible, scalable, and LLM-friendly, with support for Windows, macOS, and Ubuntu.
All the scenario components in IR-SIM, including geometry, kinematics, behaviors, sensing, contact, and maps, are declarative and represented through a human-readable YAML configuration, which can be easily modified and reused.
This declarative representation reduces the need to implement scene logic and controllers repeatedly and also exploits LLM agents' ability to read and write structured text. The provided IR-SIM agent skills further enhance this ability to reduce token usage and improve the scenario-generation success rate.
Figure~\ref{fig:scenario_gallery} illustrates diverse scenarios in IR-SIM specified through YAML configuration files. Furthermore, to enable efficient execution on accessible platforms, IR-SIM performs planar geometric queries using Shapely~\citep{GilliesShapely2025}, numerical operations using NumPy~\citep{harris2020array}, and vectorized and batched computation for world updates, particularly for LiDAR ray casting. Reusable configurations, seeded randomization, and shared evaluation protocols support navigation benchmarking, while RL interfaces connect the simulator to policy-learning libraries.
Finally, we also provide bridges to external simulators and real robotic systems to further support downstream validation through dedicated adapters.

The contributions of this work can be summarized as follows:
\begin{itemize}
    \item We develop \textbf{IR-SIM}, a lightweight declarative navigation simulator that combines rapid construction through reusable scenario components with efficient execution on accessible platforms. Generally, IR-SIM executes simulation steps faster than the compared simulators especially in large number of objects.
    \item We provide IR-SIM agent skills that guide LLM agents in constructing navigation scenarios, reducing mean authoring time by over $50\%$ compared with IR-SIM without skills while achieving approximately $95\%$ scenario validity across both evaluated models.
    \item We validate the effectiveness of IR-SIM through experiments on efficient execution, LLM-based scenario generation, and RL-based navigation policy learning and benchmarking. The code will be released as open source to facilitate future navigation research.
\end{itemize}

\section{Related Work}
\label{sec:related-work}
\subsection{Lightweight Simulators for Navigation}
Lightweight navigation simulators have a long history in robotics research. Early general-purpose examples include Stage~\citep{vaughan2008stage} and ARGoS~\citep{pinciroli2012argos}, while a more recent example is PyRoboSim~\citep{pyrobosim2026}. These simulators support multi-robot simulation in planar environments with simplified physics and rendering. More recent simulators increasingly target specific navigation tasks. For social navigation, representative examples include CrowdNav~\citep{chen2019crowdrobot}, SocialGym 2.0~\citep{chandra2024socialgym}, and SocNavBench~\citep{biswas2022socnavbench}, which focus on simulating multiple agents with social behaviors in shared spaces. RL requirements have also motivated simulators such as VMAS~\citep{bettini2022vmas} and JaxNav~\citep{rutherford2024noregrets}, which integrate with RL libraries to accelerate the training process. Finally, MVSim~\citep{blanco2023mvsim} supports mobile-robot and multi-vehicle simulation, while GPUDrive~\citep{kazemkhani2025gpudrive} and Waymax~\citep{gulino2023waymax} focus on autonomous driving. IR-SIM is a general-purpose lightweight simulator that supports a wide range of navigation tasks. Its advantages are the rapid construction of diverse, reusable scenarios and efficient execution on accessible platforms.
Table~\ref{tab:related_work_comparison} provides an overview of representative lightweight simulators in navigation research.
\begin{table}[t]
    \centering
    \caption{Capability comparison of representative navigation simulators.}
    \label{tab:related_work_comparison}
    \begingroup
    \scriptsize
    \setlength{\tabcolsep}{2pt}
    \renewcommand{\arraystretch}{1.12}
    \resizebox{\linewidth}{!}{%
    \begin{tabular}{@{}lccccccc@{}}
        \toprule
        Simulator & Lightweightness & Declarative & Kine. diversity & Geo. diversity & Sensor sim. & Contact sim. & LLM authoring \\
        \midrule
        PyRoboSim~\citep{pyrobosim2026} & High & $\checkmark$ & $\times$ (holo. only) & $\times$ (circles only) & $\checkmark$ & $\times$ & $\times$ \\
        MVSim~\citep{blanco2023mvsim} & Medium & $\checkmark$ & $\checkmark$ & $\checkmark$ & $\checkmark$ & $\checkmark$ & $\times$ \\
        VMAS~\citep{bettini2022vmas} & Medium & $\times$     & $\checkmark$ & $\checkmark$ & $\checkmark$ & $\checkmark$ & $\times$ \\
        JaxNav~\citep{rutherford2024noregrets} & Medium & $\times$     & $\times$ (diff. only) & $\checkmark$ & $\checkmark$ & $\times$ & $\times$ \\
        CrowdNav~\citep{chen2019crowdrobot} & Medium & $\times$     & $\checkmark$ & $\times$ (circles only) & $\times$ & $\times$ & $\times$ \\
        GPUDrive~\citep{kazemkhani2025gpudrive} & Low & $\checkmark$ & $\checkmark$ & $\times$ (boxes only) & $\checkmark$ & $\times$ & $\times$ \\
        Waymax~\citep{gulino2023waymax} & Low & $\times$     & $\checkmark$ & $\times$ (boxes only) & $\times$ & $\times$ & $\times$ \\
        \midrule
        \textbf{IR-SIM} (ours) & High & $\checkmark$ & $\checkmark$ & $\checkmark$ & $\checkmark$ & $\checkmark$ & $\checkmark$ \\
        \bottomrule
    \end{tabular}}
    \par\smallskip
    \begin{minipage}{\linewidth}
    \raggedright
    \textit{Lightweightness} qualitatively rates deployment burden from dependencies and installation profiles (High means lighter deployment).
    \end{minipage}
    \endgroup
    \vspace{-0.5cm}
\end{table}
\subsection{Large Language Models for Simulation}
Recent work has explored the use of LLMs in generative robotic simulation, including generating simulation tasks~\citep{wang_gensim_2024}, training supervision~\citep{wang_robogen_2024}, and three-dimensional environments~\citep{yang_holodeck_2024}, as well as adapting environment configurations for RL~\citep{zala_envgen_2024}. Another approach is for LLM agents to interact with existing software through structured tool interfaces or generated simulator-specific code. These workflows can use the Model Context Protocol (MCP)~\citep{model_context_protocol_2026} and agent skills~\citep{anthropic_agent_skills_2025}.
In this work, IR-SIM provides agent skills to guide LLM agents in constructing and modifying declarative navigation scenarios, which reduces the amount of code that must be generated and maintained and improves the success rate of scenario generation.

\subsection{Navigation Benchmarking}
Navigation benchmarking requires diverse, reproducible scenarios with customizable behaviors, together with standardized evaluation metrics~\citep{perille2020benchmarking, francis2025principles}.
Most benchmarks build on existing high-fidelity simulators for dynamic environments~\citep{nair2022dynabarn}, social navigation~\citep{biswas2022socnavbench}, and embodied navigation~\citep{qiao2026navbench}. The difficulty of constructing scenarios and the need for high-fidelity simulators limit their broader adoption. In this work, IR-SIM provides reproducible scenarios that incorporate social behavior algorithms such as Optimal Reciprocal Collision Avoidance (ORCA)~\citep{van2011reciprocal} and the Social Force Model (SFM)~\citep{helbing1995social}, together with evaluation metrics and protocols for benchmarking navigation algorithms.

\section{Simulator Design}
\label{sec:methodology}

\subsection{System Overview}
\begin{figure}[t]
    \centering
    \includegraphics[width=0.90\linewidth]{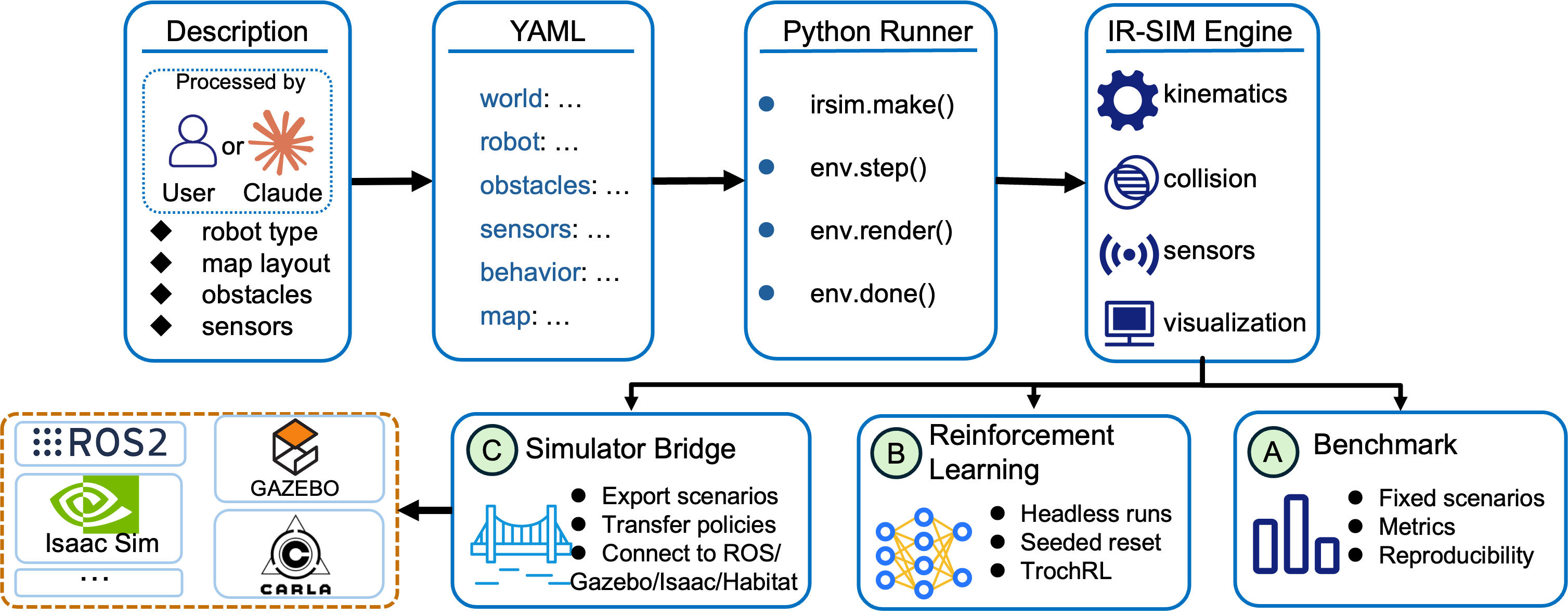}
    \caption[System overview of IR-SIM workflows]{System overview of IR-SIM and skill-native workflows.}
    \label{fig:system_overview}
    \vspace{-0.4cm}
\end{figure}

Fig.~\ref{fig:system_overview} summarizes the IR-SIM workflow from scenario description to downstream use.
The left side shows scenario authoring.
A user can either write the YAML file directly according to their needs or provide a natural-language scenario prompt, which is translated by an LLM agent, such as Codex~\citep{openai_codex_cli_2026} or Claude Code~\citep{anthropic_claude_code_2026}, into a structured YAML scenario.
The Python runner then parses the YAML file and instantiates the world, robots, behaviors, and sensors in the simulation loop.
The physics engine next advances the simulation, and the Python library Matplotlib visualizes the scenario in real time.
This design decouples scenario description, physics updates, and rendering, resulting in a stable simulator runner and flexible deployment options.
The core simulator modules are implemented in Python, allowing IR-SIM to run directly across platforms including Windows, macOS, and Ubuntu.
Finally, the Python API exposes functions for accepting control inputs, extracting state and sensor information, and generating demonstration files.

The bottom of Fig.~\ref{fig:system_overview} illustrates the main downstream applications, including navigation learning, benchmarking, and bridging to external simulators or real-world robots. To support RL-based navigation approaches, IR-SIM provides a Gym~\citep{brockman2016openai} interface and wrappers for RL libraries such as TorchRL~\citep{bou2024torchrl}. Second, scenario artifacts with seeded randomization and built-in behavior configurations enable IR-SIM to run reproducible benchmarks and evaluate different navigation algorithms across diverse scenario families.
Finally, users can use the flexible API to transfer scenarios to external simulators, e.g., Isaac Sim~\citep{NVIDIA_Isaac_Sim}, CARLA~\citep{dosovitskiy_carla_2017}, and Habitat-Sim~\citep{savva_habitat_2019}, where the algorithm can be further validated in high-fidelity simulation and realistic environments. Details of these extensions can be found in Sec.~\ref{sec:bridges-interfaces}.

\subsection{Simulation Engine}
\textbf{Physics engine}:
The physics engine of IR-SIM includes three main components: collision detection, kinematic state updates, and optional contact dynamics, which are executed at each time step. To achieve efficient execution that can run on accessible platforms, we have two main design choices. First, we only use two libraries to implement: Shapely~\citep{GilliesShapely2025} for geometric queries and NumPy~\citep{harris2020array} for numerical operations. Specifically, Shapely is a widely used open-source Python package built on the GEOS geometry library for manipulating and analyzing planar geometric objects. Second, for the most computationally expensive LiDAR simulation, we implement a batched ray-casting by using NumPy and caches static geometry to reduce repeated geometric computation. These two features bring the strong scalability of IR-SIM, which can run hundreds of robots in real time on a single CPU core. Section~\ref{sec:efficiency} evaluates the execution cost of the resulting implementation.

The kinematic state update, implemented in NumPy, advances objects in discrete time according to their configured kinematic models, using either user-specified control inputs or actions generated by built-in behaviors. Commonly used kinematic models for mobile robots are supported, including differential drive, omnidirectional, omnidirectional angular, and Ackermann. The optional contact dynamics module is also implemented in NumPy and resolves planar contacts using position-based constraint projection~\citep{mueller2007position,mueller2020rigid}.
It corrects overlaps using translations and rotations weighted by mass and inertia, updates object velocities, and estimates contact forces while accounting for friction. This allows for more realistic simulations of scenarios where robots or agents may collide or interact with each other or with obstacles in the environment.

\textbf{Rendering engine}:
Following the lightweight design principle, IR-SIM implements default rendering with the Python library Matplotlib~\citep{Hunter:2007} for simple 2D visualization.
Additionally, keyboard and mouse controls are implemented through Matplotlib's event-handling system, allowing users to interact with the simulation in real time.
For applications requiring high-fidelity rendering, such as visual sensor simulation, IR-SIM provides a bridge to external simulators, such as Isaac Sim~\citep{NVIDIA_Isaac_Sim}, CARLA~\citep{dosovitskiy_carla_2017}, and Habitat-Sim~\citep{savva_habitat_2019}.

\subsection{Scenario Representation}
\textbf{YAML configuration}:
In IR-SIM, to make scenarios easy to read and modify, all the components of a scenario are represented in a single human-readable YAML configuration file, which is parsed by the simulator to instantiate the world, robots, obstacles, behaviors, and sensors. The main YAML blocks are \texttt{world}, \texttt{robot}, \texttt{obstacle}, and \texttt{gui}; object configurations compose built-in geometry, kinematics, \texttt{behavior}, and \texttt{sensors}, while custom controllers and task logic can be implemented in Python. Each block contains relevant parameters and settings for the corresponding component. Additionally, to make the scenario more flexible and diverse, IR-SIM supports map specifications from images, where the map image is transformed into an occupancy grid map. Black pixels represent occupied cells and white pixels represent free cells.
Such a grid map can be used for the development and evaluation of path planning algorithms, such as A* search~\citep{hart1968formal} and Rapidly-exploring Random Trees (RRT)~\citep{lavalle1998rapidly}.

\textbf{Behavior modules}:
A scenario should not only contain robots and obstacles but also specify how they move and interact.
IR-SIM provides behavior modules that define motion policies for each object or group, and the behavior parameters can also be configured directly through the YAML configuration file.
The built-in social behaviors include Reciprocal Velocity Obstacles (RVO)~\citep{vandenberg2008reciprocal}, ORCA~\citep{van2011reciprocal}, SFM~\citep{helbing1995social}, and a simple goal-directed behavior called \texttt{dash}, which are commonly used in social navigation research to simulate pedestrian or dynamic agent behavior.
To support diverse behaviors, IR-SIM also provides a registry that lets users define custom behaviors according to their needs and reference them in the YAML file without changing the simulator code interface.

\subsection{LLM Authoring and Agent Skills}
LLMs can generate code and follow natural-language instructions, which reduces the effort of scenario construction in IR-SIM and other similar tools.
However, two challenges limit an unaided agent in IR-SIM: the API surface, which it may know only partly, and the runtime semantics of the built-in behaviors, which are not visible in the YAML. The agent otherwise learns these semantics only by running the scene.
The IR-SIM agent skills address both challenges through a versioned package of task-focused skills for coding agents that supplies the verified API together with tested behavior parameters.

The IR-SIM package contains five skills.
\texttt{irsim-scenario} authors, validates, and modifies YAML scenes through the blocks, it follows a YAML-first rule and carries reference sheets for the schema.
\texttt{irsim-runner} handles the execution boundary through the simple Python loop. It also provides a verified table of accessors for reading the scene state and runner templates, including one that records every pose, the static geometry, and the sensor parameters in a trace file.
\texttt{irsim-algorithm} exposes the path planners and the registration of custom behaviors and kinematics; \texttt{irsim-benchmark} structures fair comparisons through shared scenarios, seeds, episode counts, horizons, robot densities, and canonical metrics, and \texttt{irsim-dev} supports development of the library itself, including tests, linting, documentation, and releases.

\begin{figure}[t]
    \centering
    \includegraphics[width=0.95\linewidth]{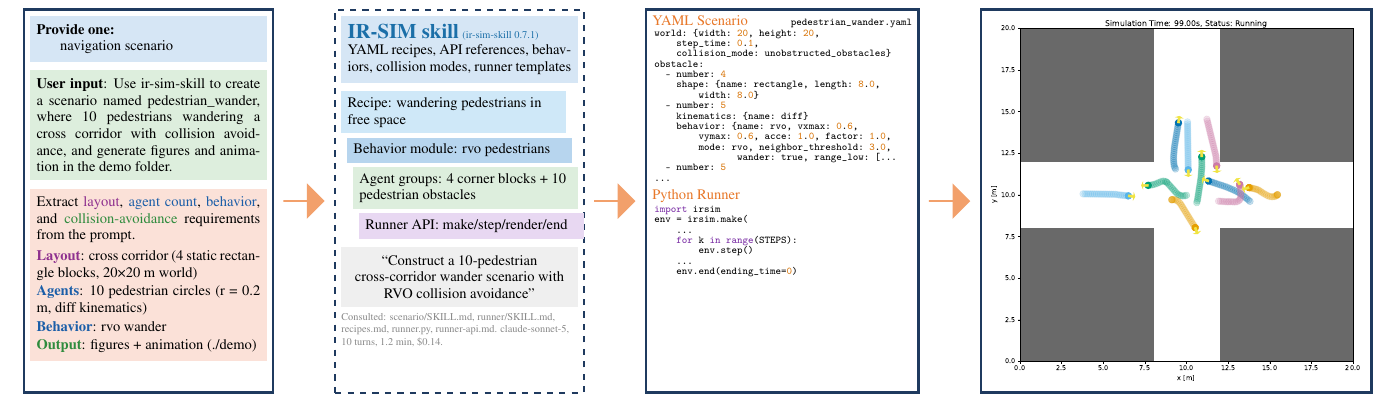}
    \caption[Construction of a pedestrian wander scenario]{Construction of the \texttt{pedestrian wander} scenario.
    The pipeline contains four stages: (a) prompting and extracting scenario elements, (b) retrieving IR-SIM skill templates and API patterns, (c) assembling YAML and Python artifacts, and (d) rendering rollout snapshots and saving the demonstration animation.}
    \label{fig:skill_pedestrian_wander}
    \figpanellabel{fig:skill_pedestrian_wander_prompt}
    \figpanellabel{fig:skill_pedestrian_wander_retrieval}
    \figpanellabel{fig:skill_pedestrian_wander_assembly}
    \figpanellabel{fig:skill_pedestrian_wander_render}
    \vspace{-0.5cm}
\end{figure}

The workflow from natural language to a scenario is illustrated in Fig.~\ref{fig:skill_pedestrian_wander}.
The user provides the description shown in Fig.~\ref{fig:skill_pedestrian_wander_prompt}; the scenario skill extracts the scene layout, the number of robots, the behavior type, and the collision-avoidance requirement, retrieves the matching recipe and reference entries, and assembles the YAML scene and the Python runner; the runner skill renders rollout snapshots and saves the requested animation.
The pipeline takes a few minutes from the prompt to the animated scenario, and the scene can be modified with a further prompt, after which the YAML file and the runner are updated.
Section~\ref{sec:construction} quantifies the effect of the skills on agent-assisted scenario construction compared with other lightweight simulators in terms of scenario validity, execution success, and agent time.
\section{Experiments}
\label{sec:experiment}
In this section, we conduct experiments to evaluate the performance of IR-SIM by addressing the following questions:
(1) How efficiently can IR-SIM execute navigation scenarios?
(2) Can LLM agents reliably construct navigation scenarios in IR-SIM, and can the provided agent skills further reduce the construction time and improve the generation success rate?
(3) Can IR-SIM support navigation benchmarking and learning with randomized and reproducible evaluation?
All the experiments are conducted on a desktop computer with an AMD Ryzen~9 7950X3D CPU and an NVIDIA GeForce RTX~4090 GPU with 24\,GB of memory.

\subsection{Execution Efficiency}
\label{sec:efficiency}
\begin{figure}[t]
    \centering
    \includegraphics[width=0.9\linewidth]{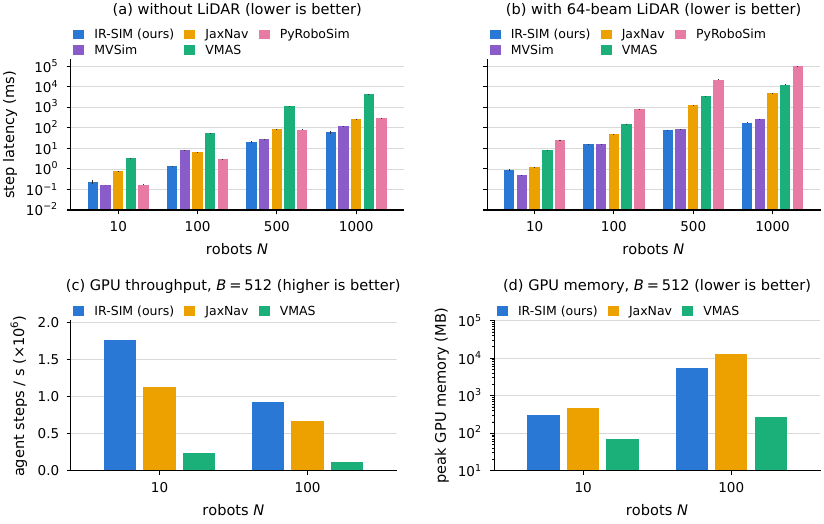}
    \caption{Execution efficiency on the shared scenario. (a),~(b): single-core CPU step latency without and with a 64-beam LiDAR for $N$ robots. (c): batched throughput on an RTX~4090 with $B=512$ worlds per call; (d): peak GPU memory of the same runs.}
    \label{fig:efficiency_bars}\label{fig:efficiency_batch}
    \vspace{-0.5cm}
\end{figure}

We evaluate the execution efficiency of IR-SIM against four recent lightweight simulators: VMAS~\citep{bettini2022vmas}, JaxNav~\citep{rutherford2024noregrets}, PyRoboSim~\citep{pyrobosim2026}, and MVSim~\citep{blanco2023mvsim}, on one shared scenario; MVSim, which cannot be stepped from Python, is timed through its C++ API.
The scenarios include $N\in\{10,100,500,1000\}$ differential-drive robots that replay the same recorded actions in a walled arena with unit-sized obstacles ($20\,\mathrm{m}\times20\,\mathrm{m}$ up to $N=100$, enlarged at constant density beyond). We run each scenario once with a $64$-beam LiDAR on every robot, for which LiDAR simulation is computationally intensive, and once without sensors.
Each simulation step comprises kinematic integration, collision handling, sensing, and observation construction. We evaluate the step latency of a single world on a CPU core and the batched throughput of many worlds on a GPU, following the agent steps per second (ASPS) protocol of GPUDrive~\citep{kazemkhani2025gpudrive}. The results are shown in Figure~\ref{fig:efficiency_bars}.

For step latency in a single world, as shown in Figure~\ref{fig:efficiency_bars} (a)(b), IR-SIM has the lowest LiDAR step latency for large teams, with $178\,\mathrm{ms}$ for $1000$ robots ($1.6\times$ faster than MVSim, $27\times$ faster than JaxNav, $71\times$ faster than VMAS). It is on par with MVSim at $100$ robots, and the compiled C++ engine of MVSim is faster for smaller teams. Key to this performance is the batched LiDAR scan, implemented in pure NumPy, the complete step latency with LiDAR grows only $11\times$ when the number of robots grows from $100$ to $1000$, a smaller increase than for MVSim ($17\times$), and others. Without LiDAR sensors, PyRoboSim, which also uses Shapely for collision detection, and the compiled MVSim perform comparably to IR-SIM with $10$ robots, but IR-SIM achieves the lowest step latency of $59\,\mathrm{ms}$ for $1000$ robots and scales better than others from $100$ to $1000$ robots, owing to its vectorized collision detection and kinematic integration in NumPy. With the optional Numba-compiled LiDAR kernel~\citep{lam2015numba}, IR-SIM further reduces the step latency with LiDAR to $3.5\,\mathrm{ms}$ for $100$ robots and $79\,\mathrm{ms}$ for $1000$ robots, $4.8\times$ and $2.2\times$ faster than the NumPy version (Appendix~\ref{app:efficiency_full}).

We also evaluate the batched throughput of 512 worlds on a GPU and test how many agent steps can be processed in $1\,\mathrm{s}$, compared with JaxNav and VMAS, which are designed for GPU execution (PyRoboSim and MVSim have no GPU mode).
Specifically, the GPU version of IR-SIM is implemented in PyTorch.
As shown in Figure~\ref{fig:efficiency_bars} (c)(d), IR-SIM achieves the highest ASPS: $1.76$ and $0.93$ million for $10$ and $100$ robots per world, respectively, against $1.13$ and $0.67$ million for JaxNav and $0.23$ and $0.11$ million for VMAS, with LiDAR at every step. IR-SIM has the second-lowest device memory usage, $5{,}364\,\mathrm{MB}$ at 512 worlds per call with $100$ robots per world, while JaxNav and VMAS use $12{,}701\,\mathrm{MB}$ and $277\,\mathrm{MB}$, respectively. In summary, IR-SIM is the most efficient simulator for large-scale multi-robot scenarios, with the lowest step latency and highest batched throughput, while maintaining a low memory footprint, owing to its NumPy- and Shapely-based implementation and batched LiDAR scanning. Further details about the scenario, timing boundaries, and step budgets are provided in Appendix~\ref{app:efficiency_full}.

\subsection{Scenario Construction}
\label{sec:scenario_construction}

\begin{figure}[t]
\centering
\includegraphics[width=0.9\linewidth]{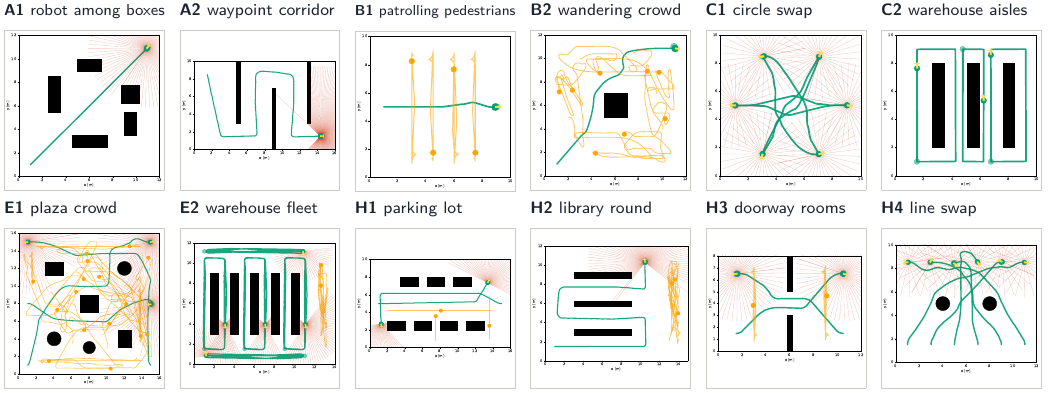}
\caption{Twelve of the thirteen construction briefs and the scenes the IR-SIM + skills agent built from them in the single-shot protocol.}
\label{fig:construction_examples}
\vspace{-0.5cm}
\end{figure}

\newcommand{\CONSTfirstIRSIMoneshotopus}{87}
\newcommand{\CONSTfirstIRSIMmaxoneshotopus}{95}
\newcommand{\CONSTfirstIRSIMrangeoneshotopus}{87$ to $95}
\newcommand{\CONSTyamlIRSIMoneshotopus}{52}
\newcommand{\CONSTpyIRSIMoneshotopus}{129}
\newcommand{\CONSTpyVMASoneshotopus}{316}
\newcommand{\CONSTpyPRSoneshotopus}{293}
\newcommand{\CONSTtIRSIMoneshotopus}{1.2}
\newcommand{\CONSTtVMASoneshotopus}{3.7}
\newcommand{\CONSTnattemptsoneshotopus}{39}
\newcommand{\CONSTfirstVMASoneshotopus}{97}
\newcommand{\CONSTfirstPRSoneshotopus}{90}
\newcommand{\CONSTexecIRSIMoneshotopus}{90}
\newcommand{\CONSTaccIRSIMoneshotopus}{93}
\newcommand{\CONSTaccVMASoneshotopus}{97}
\newcommand{\CONSTaccPRSoneshotopus}{99}
\newcommand{\CONSTfirstSKILLoneshotopus}{95}
\newcommand{\CONSTfirstPLAINoneshotopus}{87}
\newcommand{\CONSTcostSKILLoneshotopus}{0.51}
\newcommand{\CONSTtokSKILLoneshotopus}{0.3}
\newcommand{\CONSTcostconSKILLoneshotopus}{0.51}
\newcommand{\CONSTcostPLAINoneshotopus}{1.04}
\newcommand{\CONSTtokPLAINoneshotopus}{0.9}
\newcommand{\CONSTcostconPLAINoneshotopus}{1.04}
\newcommand{\CONSTcostVMASoneshotopus}{1.02}
\newcommand{\CONSTtokVMASoneshotopus}{0.7}
\newcommand{\CONSTcostconVMASoneshotopus}{1.02}
\newcommand{\CONSTcostPRSoneshotopus}{1.00}
\newcommand{\CONSTtokPRSoneshotopus}{0.8}
\newcommand{\CONSTcostconPRSoneshotopus}{1.00}
\newcommand{\CONSTcostJAXoneshotopus}{1.18}
\newcommand{\CONSTtokJAXoneshotopus}{0.5}
\newcommand{\CONSTcostconJAXoneshotopus}{1.18}
\newcommand{\CONSTfirstJAXoneshotopus}{89}
\newcommand{\CONSTnJAXoneshotopus}{18}
\newcommand{\CONSTpyJAXoneshotopus}{258}
\newcommand{\CONSTtJAXoneshotopus}{3.5}
\newcommand{\CONSThumanrowsall}{%
\textbf{IR-SIM} + Humans$^{\ddagger}$ ($n=15$) & 100.0 & 60.0 & 88.7 & -- & 828 & -- & 64 & 34 \\
}
\newcommand{\CONSThumanrowscon}{%
\textbf{IR-SIM} + Humans$^{\ddagger}$ ($n=5$) & 100.0 & 60.0 & 91.7 & -- & 1194 & -- & 69 & 34 \\
}
\newcommand{\CONSThumanrowsmod}{%
\textbf{IR-SIM} + Humans$^{\ddagger}$ ($n=10$) & 100.0 & 60.0 & 87.2 & -- & 646 & -- & 62 & 34 \\
}
\label{sec:construction}
In this experiment, we measure how reliably and quickly an LLM agent builds navigation scenarios using IR-SIM, compare its performance with that on other lightweight simulators, and evaluate the effect of IR-SIM's agent skills described in Section~\ref{sec:methodology}.
Thirteen simulator-independent construction briefs span five families: (A) static obstacles, (B) pedestrian encounters, (C) multi-robot navigation, (E) rich scenes with many robots and pedestrians moving according to standard behaviors, seeded crowds and layouts, and sensors with a limited field of view, and (H) further mixed scenes; all thirteen lie in the capability intersection of IR-SIM, VMAS, and PyRoboSim (JaxNav is evaluated on six compatible briefs), require no task-specific control code, and were solved by hand beforehand (Appendix~\ref{app:construction_briefs}). Figure~\ref{fig:construction_examples} shows twelve construction briefs with the scenes built from them.

We use Claude Code with \texttt{claude-opus-5-5} to receive the brief and write the deliverable once without executing anything. The agent attempts every supported brief-condition pair three times.
Following the staged validation of GenSim~\citep{wang_gensim_2024}, an independent validator grades each deliverable as \emph{executable} (it runs headlessly and writes a rollout trace) and \emph{valid} (all acceptance checks of the brief pass, inspected through the simulator's API and the recorded rollout). Specification accuracy is the mean fraction of acceptance checks passed.
Turns, Time, Cost, and Lines report means over all attempts: documentation and file-tool turns, agent wall-clock time excluding validation, API list-price cost, and non-comment Python/YAML lines, respectively.
\begin{table}[t]
\caption{Single-shot scenario construction with \texttt{claude-opus-5-5} and \texttt{claude-sonnet-5}. $^{\dagger}$JaxNav covers A1, A2, C1, C2, E3 and H4 only (18 attempts). $^{\ddagger}$Human IR-SIM submissions on the same briefs, graded with the same validator.}
\label{tab:construction_oneshot_opus55}
\centering
\small
\setlength{\tabcolsep}{3.5pt}
\begin{tabular}{lcccccccc}
\toprule
\multirow{2}{*}{Simulator} & Executable & Valid & Spec.\ acc. & \multirow{2}{*}{Turns} & Time & Cost & \multicolumn{2}{c}{Lines} \\
\cmidrule(lr){8-9}
 & (\%) & (\%) & (\%) & & (s) & (USD) & Python & YAML \\
\midrule
\multicolumn{9}{l}{\textit{\texttt{claude-opus-5-5}: construction briefs A1, A2, B1, B2, C1, C2, E1, E2, E3, H1, H2, H3, H4}} \\
\textbf{IR-SIM} + skills & \textbf{100.0} & 94.9 & \textbf{99.5} & \textbf{11.4} & \textbf{98} & \textbf{0.51} & \textbf{78} & \textbf{35} \\
\textbf{IR-SIM} & 89.7 & 87.2 & 87.2 & 24.3 & 204 & 1.04 & 179 & 68 \\
VMAS & 97.4 & \textbf{97.4} & 97.4 & 23.2 & 238 & 1.02 & 316 & 0 \\
PyRoboSim & \textbf{100.0} & 89.7 & 99.1 & 24.4 & 212 & 1.00 & 293 & 9 \\
JaxNav$^{\dagger}$ & 88.9 & 88.9 & 88.9 & 12.1 & 229 & 1.18 & 258 & 0 \\
\midrule
\multicolumn{9}{l}{\textit{\texttt{claude-sonnet-5}: construction briefs A1, A2, B1, B2, C1, C2, E1, E2, E3, H1, H2, H3, H4}} \\
\textbf{IR-SIM} + skills & \textbf{100.0} & \textbf{94.9} & \textbf{99.7} & \textbf{9.0} & \textbf{41} & \textbf{0.17} & \textbf{64} & \textbf{41} \\
\textbf{IR-SIM} & 94.9 & 61.5 & 92.7 & 15.8 & 92 & 0.25 & 126 & 58 \\
VMAS & 71.8 & 64.1 & 71.3 & 19.1 & 137 & 0.34 & 222 & 0 \\
PyRoboSim & \textbf{100.0} & 87.2 & 98.8 & 22.2 & 127 & 0.38 & 210 & 8 \\
JaxNav$^{\dagger}$ & 72.2 & 72.2 & 72.2 & 13.6 & 172 & 0.55 & 206 & 0 \\
\ifdefined\CONSThumanrowscon
\midrule
\multicolumn{9}{l}{\textit{Humans: construction brief A1}} \\
\CONSThumanrowscon
\fi
\bottomrule
\end{tabular}
\vspace{-0.5cm}
\end{table}

The results are listed in Table~\ref{tab:construction_oneshot_opus55}. IR-SIM with skills achieves the lowest generation time, API cost, and deliverable size on the construction briefs, with $95\%$ validity.
On these briefs, skills reduce mean agent time from $204$ to $98\,\mathrm{s}$ and cost from \$$1.04$ to \$$0.51$ per attempt, while validity rises from $87\%$ to $95\%$. Compared to human authors on A1, the agent is $19.2\times$ faster, while achieving higher validity ($3/3$ versus $3/5$).
With the smaller \texttt{claude-sonnet-5} (second block of Table~\ref{tab:construction_oneshot_opus55}), the conditions also differ in validity: IR-SIM with skills is valid in $37$ of $39$ attempts at $41\,\mathrm{s}$ and \$$0.17$, against $62\%$ for plain IR-SIM, $64\%$ for VMAS, $87\%$ for PyRoboSim, and $72\%$ for JaxNav, at $92$ to $172\,\mathrm{s}$ per attempt; attempts with code-generating interfaces fail because of runtime errors and controllers that miss goals, whereas the skills keep the declarative scene correct. IR-SIM with skills achieves the shortest generation time, lowest cost, and fewest lines of code among the compared agent conditions, with approximately $95\%$ validity across both evaluated models and the highest validity with smaller model \texttt{claude-sonnet-5}.
Detailed test results are in Appendix~\ref{app:construction_results}.

\subsection{Navigation Benchmarking and Learning}
\label{sec:social_benchmark}
\textbf{Multi-robot navigation benchmarking:} We evaluate IR-SIM's support for reproducible navigation benchmarking by comparing eight reactive, optimization-based, and learning-based controllers under a shared scenario specification and evaluation protocol: ORCA~\citep{van2011reciprocal}, VO-polytope~\citep{huang2023velocity}, AVOCADO~\citep{martinez2025avocado}, the RDA planner~\citep{han2023rda}, and the learned policies SARL~\citep{chen2019crowdrobot}, CADRL~\citep{chen2017decentralized}, LSTM-RL~\citep{everett2018motion}, and RL-RVO~\citep{han2022reinforcement}.
The task comprises $5$ or $10$ differential-drive robots with random convex polygonal footprints in a $10\,\mathrm{m}\times10\,\mathrm{m}$ world. Episode success requires every robot to reach its goal without collision within $500$ steps. The results are listed in Table~\ref{tab:social_benchmark}, which reports episode-level success, collision, and timeout rates (\%) and navigation time of robots that reach their goals (mean $\pm$ standard deviation) for $100$ randomized episodes per setting with a fixed seed. Specially, the CADRL, SARL, and LSTM-RL policies focus on social navigation and thus has lower success rates than the other multi-robot controllers, which align with our intuition.
The experiment demonstrates that IR-SIM can combine shared scenario generation and metric collection to benchmark heterogeneous navigation controllers reproducibly.
Controller settings, geometry representations, and policy provenance are detailed in Appendix~\ref{app:social_benchmark}.

\begin{table}[t]
\caption[Multi-robot navigation benchmark]{Example IR-SIM benchmark for multi-robot collision avoidance.}
\label{tab:social_benchmark}
\centering
\scriptsize
\setlength{\tabcolsep}{3pt}
\renewcommand{\arraystretch}{0.95}
\resizebox{0.98\linewidth}{!}{
\begin{tabular}{l|cccc|cccc}
\toprule
Method & \multicolumn{4}{c}{5 robots} & \multicolumn{4}{c}{10 robots} \\
\cmidrule(lr){2-5} \cmidrule(lr){6-9}
 & Succ.$\uparrow$ & Coll.$\downarrow$ & Tout.$\downarrow$ & Time (s)$\downarrow$ & Succ.$\uparrow$ & Coll.$\downarrow$ & Tout.$\downarrow$ & Time (s)$\downarrow$ \\
\midrule
ORCA~\citep{van2011reciprocal} & \textbf{100} & \textbf{0} & \textbf{0} & $5.05\pm2.47$ & \textbf{90} & 1 & 9 & $5.41\pm3.04$ \\
VO-polytope~\citep{huang2023velocity} & 83 & 17 & \textbf{0} & $\mathbf{4.74}\pm2.28$ & 45 & 55 & \textbf{0} & $\mathbf{4.31}\pm2.44$ \\
AVOCADO~\citep{martinez2025avocado} & \textbf{100} & \textbf{0} & \textbf{0} & $5.23\pm2.38$ & 86 & \textbf{0} & 14 & $5.50\pm2.75$ \\
\midrule
RDA~\citep{han2023rda} & 20 & \textbf{0} & 80 & $5.03\pm3.48$ & 0 & \textbf{0} & 100 & $5.94\pm6.04$ \\
\midrule
SARL~\citep{chen2019crowdrobot} & 41 & \textbf{0} & 59 & $10.90\pm7.80$ & 26 & 11 & 63 & $10.96\pm8.41$ \\
CADRL~\citep{chen2017decentralized} & 9 & 6 & 85 & $8.79\pm6.87$ & 0 & 40 & 60 & $8.93\pm7.30$ \\
LSTM-RL~\citep{everett2018motion} & 39 & 13 & 48 & $9.97\pm6.92$ & 0 & 36 & 64 & $12.69\pm11.02$ \\
RL-RVO~\citep{han2022reinforcement} & 89 & 6 & 5 & $5.82\pm3.46$ & 73 & 20 & 7 & $6.87\pm4.19$ \\
\bottomrule
\end{tabular}
}
\vspace{-0.5cm}
\end{table}

\textbf{Reinforcement learning:}
We implement the classic RL-based social navigation approach CrowdNav~\citep{chen2019crowdrobot} in our IR-SIM simulation with a TorchRL adapter to demonstrate the capability and efficiency of IR-SIM as a training backend for learning-based navigation policies in multi-agent scenarios.
Specifically, we compare 1) the original CrowdNav algorithm~\citep{chen2019crowdrobot} and 2) the original CrowdNav algorithm with IR-SIM backend and a TorchRL adapter (our framework). The results are listed in Table~\ref{tab:social_benchmark_rl}. We can see that the original CrowdNav algorithm with IR-SIM backend has $1.1$ times the training speed, which time cost is mainly on the single world simulation step. While, by using our IR-SIM and TorchRL adapter, where the world simulation is batched, the training time is reduced significantly from $6.99\,\mathrm{h}$ to $0.76\,\mathrm{h}$ ($9.2\times$ speedup), which could further be reduced to $0.51\,\mathrm{h}$ ($13.7\times$ speedup) with GPU acceleration, while achieving the same policy performance.

\begin{table}[t]
\centering
\small
\caption{Training CrowdNav's SARL policy: CrowdNav's original implementation and simulator versus IR-SIM with a batched TorchRL implementation of the same recipe.}
\label{tab:policy_training_pipeline}\label{tab:social_benchmark_rl}
\resizebox{\linewidth}{!}{%
\begin{tabular}{lllr rrrr rrrr}
\toprule
 & & & & \multicolumn{4}{c}{5 humans} & \multicolumn{4}{c}{10 humans} \\
\cmidrule(lr){5-8} \cmidrule(lr){9-12}
Method & Simulator & Hardware & Worlds & Sim.\ (ms) & Net.\ (ms) & Time (h) & Speed-up & Sim.\ (ms) & Net.\ (ms) & Time (h) & Speed-up \\
\midrule
CrowdNav & \texttt{CrowdSim} & CPU & 1 & 6.86 & 27.77 & 6.99 & 1.0$\times$ & 31.17 & 29.43 & 13.11 & 1.0$\times$ \\
CrowdNav & IR-SIM & CPU & 1 & 2.50 & 27.00 & 6.13 & 1.1$\times$ & 4.01 & 28.17 & 7.78 & 1.7$\times$ \\
Batched IR-SIM + TorchRL & IR-SIM & CPU & 16 & 0.43 & 0.47 & 0.76 & 9.2$\times$ & 0.72 & 0.76 & 1.27 & 10.3$\times$ \\
Batched IR-SIM + TorchRL & IR-SIM & CPU + GPU & 16 & \textbf{0.41} & \textbf{0.11} & \textbf{0.51} & \textbf{13.7}$\times$ & \textbf{0.70} & \textbf{0.12} & \textbf{0.70} & \textbf{18.8}$\times$ \\
\bottomrule
\end{tabular}}
\vspace{-0.5cm}
\end{table}

\section{Conclusion and Future Work}
\label{sec:conclusion}
\textbf{Conclusion:} We presented IR-SIM, a lightweight declarative simulator for navigation learning and benchmarking, which enables users to rapidly construct diverse, reusable navigation scenarios and execute them efficiently on accessible platforms.
Experiments demonstrate faster execution than the compared lightweight simulators in the evaluated LiDAR scenarios.
With Claude agents, the provided agent skills reduce scenario-generation time and improve scenario validity compared with IR-SIM without skills.
We further demonstrate that IR-SIM supports reproducible navigation benchmarking and that using IR-SIM with TorchRL reduces the training time for SARL, a classic RL-based social navigation approach, while maintaining the baseline success rate in the evaluated setting.
We will open-source the simulator to enable further research.

\textbf{Future work:} IR-SIM is currently limited to 2D navigation and does not directly support 3D navigation or general manipulation tasks.
It also lacks the complex robot dynamics required for legged locomotion and aerial vehicles, which limits its applicability to these settings.
Future work will focus on extending IR-SIM to 3D navigation and more complex tasks, with richer robot dynamics and behaviors.

\subsection*{AI use statement}
An AI assistant was used to polish the writing of the abstract and introduction, and to check for grammar and spelling errors in the main text.
AI assistants also provided feedback on experimental design and methodology, assisted with the implementation and debugging of experiment and analysis code, and supported the interpretation of experimental results.

\subsection*{Reproducibility statement}
Appendix~\ref{app:model_equations} provides the kinematic and sensor models. Appendix~\ref{app:runtime_learning} documents the CPU and GPU execution measurements, and Appendix~\ref{app:construction_full} provides the scenario-construction protocol, validation checks, and complete results.
Appendix~\ref{app:social_benchmark} lists the scenario, evaluation protocol, and controller settings of the social navigation benchmark.
We plan to release the core IR-SIM codebase, TorchRL wrappers, and bridges to external simulators as open source.

\bibliography{references}
\bibliographystyle{iclr2027_conference}

\clearpage
\appendix
\section*{Appendix}

\setcounter{tocdepth}{2}
\tableofcontents

\newcommand{\appscenariofigurepair}[7]{%
\begin{figure}[p]
\centering
\includegraphics[width=0.86\linewidth]{#1}
\par\smallskip
{\scriptsize #2}
\par\medskip
\includegraphics[width=0.86\linewidth]{#3}
\par\smallskip
{\scriptsize #4}
\caption{#5}
\label{#6}
\label{#7}
\end{figure}
}

\section{Simulator Models and Implementation Details}
\label{app:model_equations}
This section summarizes representative kinematic and LiDAR models used by IR-SIM.
Let $\Delta t$ denote the simulation step time.
The planar robot pose is written as $\mathbf{x}=[x,y,\theta]^\top$, where $(x,y)$ is the robot position and $\theta$ is its heading.
IR-SIM uses explicit discrete updates, so each object state is advanced from step $k$ to $k+1$ according to its configured kinematic model.
Differential-drive, Ackermann, omnidirectional, and omnidirectional kinematics with yaw control are selected in the YAML scenario with \texttt{diff}, \texttt{acker}, \texttt{omni}, and \texttt{omni\_angular}, respectively.

\subsection{Differential-Drive Kinematics}

For a differential-drive robot, the action is $\mathbf{u}=[v,\omega]^\top$, where $v$ is the forward speed and $\omega$ is the angular speed.
The continuous-time model is
\begin{equation}
\dot{\mathbf{x}}
=
\begin{bmatrix}
\dot{x}\\
\dot{y}\\
\dot{\theta}
\end{bmatrix}
=
\begin{bmatrix}
v\cos\theta\\
v\sin\theta\\
\omega
\end{bmatrix}.
\label{eq:appendix_diff_continuous}
\end{equation}
IR-SIM applies an Euler update,
\begin{equation}
\mathbf{x}_{k+1}
=
\mathbf{x}_{k}
+
\Delta t
\begin{bmatrix}
\cos\theta_k & 0\\
\sin\theta_k & 0\\
0 & 1
\end{bmatrix}
\begin{bmatrix}
v_k\\
\omega_k
\end{bmatrix},
\quad
\theta_{k+1}\leftarrow \operatorname{wrap}_{\pi}(\theta_{k+1}),
\label{eq:appendix_diff_discrete}
\end{equation}
where $\operatorname{wrap}_{\pi}(\cdot)$ wraps the heading angle to the range $(-\pi, \pi]$.

\subsection{Ackermann Kinematics}

For an Ackermann vehicle, IR-SIM stores the state as $\mathbf{x}=[x,y,\theta,\psi]^\top$, where $\psi$ is the steering angle and $L$ is the wheelbase.
In the default steering mode, the action is $\mathbf{u}=[v,\delta]^\top$, where $\delta$ is the steering command stored for the next step.
The pose update uses the current steering angle $\psi_k$:
\begin{equation}
\begin{bmatrix}
x_{k+1}\\
y_{k+1}\\
\theta_{k+1}\\
\psi_{k+1}
\end{bmatrix}
=
\begin{bmatrix}
x_k\\
y_k\\
\theta_k\\
\psi_k
\end{bmatrix}
+
\Delta t
\begin{bmatrix}
\cos\theta_k & 0\\
\sin\theta_k & 0\\
\tan\psi_k/L & 0\\
0 & 1
\end{bmatrix}
\begin{bmatrix}
v_k\\
\delta_k
\end{bmatrix},
\quad
\theta_{k+1}\leftarrow \operatorname{wrap}_{\pi}(\theta_{k+1}).
\label{eq:appendix_ackermann_discrete}
\end{equation}

\subsection{Omnidirectional Kinematics}

IR-SIM also supports omnidirectional robots.
For the \texttt{omni} model, the action is a body-frame velocity $\mathbf{u}=[v_f,v_l]^\top$, where $v_f$ and $v_l$ are forward and lateral speeds.
The heading is preserved, and the world-frame velocity is obtained by rotating the body-frame velocity:
\begin{equation}
\begin{bmatrix}
x_{k+1}\\
y_{k+1}\\
\theta_{k+1}
\end{bmatrix}
=
\begin{bmatrix}
x_k\\
y_k\\
\theta_k
\end{bmatrix}
+
\Delta t
\begin{bmatrix}
\cos\theta_k & -\sin\theta_k\\
\sin\theta_k & \cos\theta_k\\
0 & 0
\end{bmatrix}
\begin{bmatrix}
v_{f,k}\\
v_{l,k}
\end{bmatrix}.
\label{eq:appendix_omni_discrete}
\end{equation}

\subsection{Omnidirectional Kinematics with Yaw Control}

The \texttt{omni\_angular} model extends the \texttt{omni} model with independent yaw control.
The state is $\mathbf{x}=[x,y,\theta]^\top$, and the action is a body-frame velocity $\mathbf{u}=[v_f,v_l,\omega]^\top$, where $v_f$ and $v_l$ are the forward and lateral speeds and $\omega$ is the yaw rate.
The continuous-time model is
\begin{equation}
\dot{\mathbf{x}}
=
\begin{bmatrix}
\dot{x}\\
\dot{y}\\
\dot{\theta}
\end{bmatrix}
=
\begin{bmatrix}
v_f\cos\theta - v_l\sin\theta\\
v_f\sin\theta + v_l\cos\theta\\
\omega
\end{bmatrix}.
\label{eq:appendix_omni_angular_continuous}
\end{equation}
IR-SIM applies an Euler update,
\begin{equation}
\mathbf{x}_{k+1}
=
\mathbf{x}_{k}
+
\Delta t
\begin{bmatrix}
\cos\theta_k & -\sin\theta_k & 0\\
\sin\theta_k & \cos\theta_k & 0\\
0 & 0 & 1
\end{bmatrix}
\begin{bmatrix}
v_{f,k}\\
v_{l,k}\\
\omega_k
\end{bmatrix},
\quad
\theta_{k+1}\leftarrow \operatorname{wrap}_{\pi}(\theta_{k+1}).
\label{eq:appendix_omni_angular_discrete}
\end{equation}
The translation is rotated by the current heading $\theta_k$ before the heading itself is updated.
Unlike the \texttt{omni} model, whose heading stays fixed, the heading of an \texttt{omni\_angular} robot is controlled independently of its travel direction, so the robot can translate in any direction while rotating.

\subsection{Velocity Noise Model}

When velocity noise is enabled in the YAML scenario, IR-SIM perturbs the commanded velocity before applying the kinematic update.
For differential-drive and Ackermann models, the noisy command is
\begin{equation}
\tilde{\mathbf{u}}
=
\mathbf{u}
+
\boldsymbol{\epsilon},
\quad
\boldsymbol{\epsilon}
\sim
\mathcal{N}\!\left(
\mathbf{0},
\begin{bmatrix}
\alpha_1 v^2+\alpha_2 u_2^2 & 0\\
0 & \alpha_3 v^2+\alpha_4 u_2^2
\end{bmatrix}
\right),
\label{eq:appendix_velocity_noise}
\end{equation}
where $u_2$ denotes the second action component, either angular speed or steering command depending on the kinematic model.
For the \texttt{omni} and \texttt{omni\_angular} models, IR-SIM applies independent zero-mean Gaussian noise to each body-frame velocity channel, with a constant variance that does not scale with the command:
$\boldsymbol{\epsilon}\sim\mathcal{N}\big(\mathbf{0},\operatorname{diag}(\alpha_f,\alpha_l)\big)$ for \texttt{omni} and $\boldsymbol{\epsilon}\sim\mathcal{N}\big(\mathbf{0},\operatorname{diag}(\alpha_f,\alpha_l,\alpha_\omega)\big)$ for \texttt{omni\_angular}.

\subsection{LiDAR Simulation and Collision Checking}

For a 2D LiDAR with $N$ beams and angular range $\phi$, beam $i$ has relative angle
\begin{equation}
\beta_i = -\frac{\phi}{2} + i\Delta \beta,
\quad
\Delta \beta = \frac{\phi}{N-1},
\quad
i=0,\dots,N-1.
\label{eq:appendix_lidar_angle}
\end{equation}
Given the sensor origin $\mathbf{o}$ and heading $\theta_s$, beam $i$ is the ray
\begin{equation}
\mathbf{p}_i(r)
=
\mathbf{o}
+
r
\begin{bmatrix}
\cos(\theta_s+\beta_i)\\
\sin(\theta_s+\beta_i)
\end{bmatrix},
\quad
0<r\le r_{\max},
\label{eq:appendix_lidar_ray}
\end{equation}
and the reported range is the distance to its nearest intersection with the obstacle boundaries,
\begin{equation}
r_i
=
\min\Big(r_{\max},\;\min_{r\in\mathcal{I}_i} r\Big),
\label{eq:appendix_lidar_range}
\end{equation}
where $\mathcal{I}_i$ is the set of ranges at which beam $i$ meets a boundary segment.
IR-SIM implements the ray casting in NumPy: at each step, the beams of all LiDAR sensors are intersected with the obstacle boundary segments in one vectorized pass.

Collision checking is performed on planar geometric footprints.
Let $\mathcal{G}_i(\mathbf{x}_i)$ denote the footprint of object $i$ transformed by its current state.
Two objects are considered in collision when
\begin{equation}
\mathcal{G}_i(\mathbf{x}_i)\cap \mathcal{G}_j(\mathbf{x}_j)\ne \emptyset.
\label{eq:appendix_collision}
\end{equation}
IR-SIM implements this test with Shapely, using a spatial index to select candidate pairs before the exact intersection test.

\section{Computational Efficiency}
\label{app:runtime_learning}

\subsection{Cross-Simulator Step-Time Measurements}
\label{app:efficiency_full}
\begin{figure}[t]
    \centering
    \includegraphics[width=0.98\linewidth]{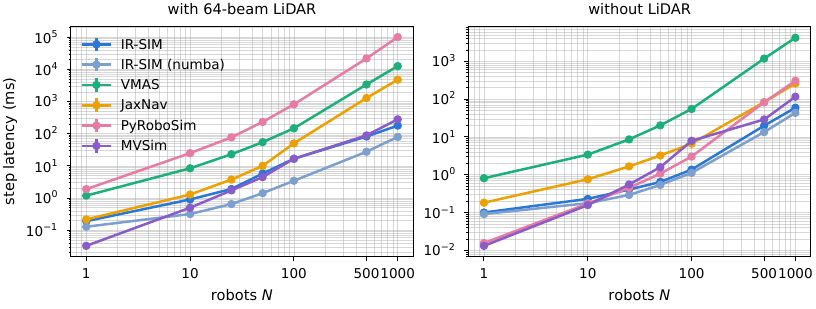}
    \caption{Single-core CPU scaling on the shared navigation scenario of Section~\ref{sec:efficiency}: mean step latency against the number of robots with a 64-beam LiDAR on every robot (left) and without LiDAR (right); error bars give the standard deviation over five process-level runs, and both axes are logarithmic.}
    \label{fig:efficiency}
\end{figure}

Tables~\ref{tab:appendix_efficiency_full} and~\ref{tab:appendix_efficiency_full_nolidar} and Figure~\ref{fig:efficiency} report the full single-core CPU measurements behind Figure~\ref{fig:efficiency_bars}, with and without LiDAR.

\paragraph{Scenario.}
All simulators receive the same obstacle geometry, initial poses, and recorded random actions.
Differential-drive disc robots of radius $0.3\,\mathrm{m}$ move with a time step of $\Delta t=0.1\,\mathrm{s}$ in a walled $20\,\mathrm{m}\times20\,\mathrm{m}$ arena with $14$ square obstacles; beyond $N=100$, the arena grows as $20\sqrt{N/100}\,\mathrm{m}$ to keep the robot density constant.
In the LiDAR condition, every robot carries a $64$-beam, $360^\circ$ LiDAR with a $6\,\mathrm{m}$ range.

\paragraph{Timing.}
One step covers kinematics, collision handling, LiDAR sensing when enabled, and observation construction.
Each configuration runs on one pinned core of an AMD Ryzen~9~7950X3D in five independent processes, timing $1000$ steps after $100$ warm-up steps; at $N\geq500$ the step budget is reduced, as marked in the tables.
MVSim is timed through its C++ API with the same step content, since its Python interface cannot advance a single step; its robots are octagons of the same circumradius, and Box2D contact forces replace the stop rule.
IR-SIM (numba) is the same IR-SIM build with its optional LiDAR kernel compiled by Numba~\citep{lam2015numba}; the main text reports the dependency-free NumPy build.

\begin{table}[t]
\caption{Single-core CPU results on the shared scenario (mean over five runs): step latency (ms) with its standard deviation and 95th percentile, agent steps per second, setup time (s), and peak resident memory (MB). Bold: best per $N$; $\uparrow$/$\downarrow$: higher/lower is better. The arena side length is $20\,\mathrm{m}$ up to $N=100$ and $45\,\mathrm{m}$ and $63\,\mathrm{m}$ at $N=500$ and $1000$. $^\dagger$: 100--200 timed steps instead of 1000. MVSim's setup excludes the Python start-up of the other simulators.}
\label{tab:appendix_efficiency_full}\label{tab:appendix_efficiency_full_nolidar}
\centering
\scriptsize
\setlength{\tabcolsep}{2.4pt}
\begin{tabular}{@{}lrrrrrrrrrrr@{}}
\toprule
\multirow{2}{*}{Simulator} & \multirow{2}{*}{$N$} & \multicolumn{5}{c}{with 64-beam LiDAR} & \multicolumn{5}{c}{without LiDAR} \\
\cmidrule(lr){3-7}\cmidrule(l){8-12}
 & & Mean\,$\downarrow$ & p95\,$\downarrow$ & Steps/s\,$\uparrow$ & Setup\,$\downarrow$ & RSS\,$\downarrow$ & Mean\,$\downarrow$ & p95\,$\downarrow$ & Steps/s\,$\uparrow$ & Setup\,$\downarrow$ & RSS\,$\downarrow$ \\
\midrule
IR-SIM & 1 & 0.19{\tiny$\pm$0.00} & 0.21 & 5237 & 0.26 & 98 & 0.10{\tiny$\pm$0.01} & 0.11 & 10067 & 0.31 & 97 \\
IR-SIM & 10 & 0.90{\tiny$\pm$0.11} & 1.10 & 11260 & 0.28 & 100 & 0.23{\tiny$\pm$0.06} & 0.34 & 45387 & 0.37 & 97 \\
IR-SIM & 25 & 1.91{\tiny$\pm$0.03} & 2.20 & 13110 & 0.33 & 103 & 0.41{\tiny$\pm$0.08} & 0.68 & 63718 & 0.44 & 98 \\
IR-SIM & 50 & 5.77{\tiny$\pm$1.49} & 6.74 & 9693 & 0.49 & 107 & 0.65{\tiny$\pm$0.01} & 0.81 & 77454 & 0.41 & 99 \\
IR-SIM & 100 & 16.6{\tiny$\pm$0.13} & 18.1 & 6014 & 0.37 & 123 & 1.37{\tiny$\pm$0.02} & 1.49 & 73122 & 0.47 & 101 \\
IR-SIM & 500$^\dagger$ & 81.8{\tiny$\pm$0.38} & 86.7 & 6112 & 0.85 & 225 & 19.9{\tiny$\pm$2.52} & 27.2 & 25478 & 0.90 & 113 \\
IR-SIM & 1000$^\dagger$ & 178{\tiny$\pm$3.35} & 193 & 5622 & 1.34 & 353 & 59.4{\tiny$\pm$9.65} & 73.3 & 17227 & 1.42 & 129 \\
\midrule
IR-SIM (numba) & 1 & 0.13{\tiny$\pm$0.00} & 0.16 & 7796 & 0.46 & 271 & 0.09{\tiny$\pm$0.01} & 0.10 & 11096 & 0.28 & 97 \\
IR-SIM (numba) & 10 & \textbf{0.32}{\tiny$\pm$0.00} & \textbf{0.43} & \textbf{31398} & 0.48 & 272 & 0.18{\tiny$\pm$0.01} & \textbf{0.26} & 56663 & 0.28 & 97 \\
IR-SIM (numba) & 25 & \textbf{0.65}{\tiny$\pm$0.02} & \textbf{0.89} & \textbf{38529} & 0.53 & 273 & \textbf{0.29}{\tiny$\pm$0.00} & \textbf{0.48} & \textbf{85666} & 0.32 & 98 \\
IR-SIM (numba) & 50 & \textbf{1.42}{\tiny$\pm$0.01} & \textbf{1.59} & \textbf{35296} & 0.65 & 275 & \textbf{0.54}{\tiny$\pm$0.01} & \textbf{0.68} & \textbf{92893} & 0.35 & 99 \\
IR-SIM (numba) & 100 & \textbf{3.45}{\tiny$\pm$0.02} & \textbf{3.59} & \textbf{29025} & 0.64 & 279 & \textbf{1.12}{\tiny$\pm$0.03} & \textbf{1.29} & \textbf{88956} & 0.44 & 101 \\
IR-SIM (numba) & 500$^\dagger$ & \textbf{27.5}{\tiny$\pm$0.20} & \textbf{31.9} & \textbf{18186} & 1.14 & 307 & \textbf{13.4}{\tiny$\pm$0.12} & \textbf{17.6} & \textbf{37208} & 0.64 & 113 \\
IR-SIM (numba) & 1000$^\dagger$ & \textbf{79.1}{\tiny$\pm$0.85} & \textbf{86.9} & \textbf{12643} & 1.83 & 342 & \textbf{43.5}{\tiny$\pm$0.33} & \textbf{53.1} & \textbf{22968} & 1.05 & 129 \\
\midrule
VMAS & 1 & 1.19{\tiny$\pm$0.00} & 1.21 & 841 & 0.06 & 536 & 0.81{\tiny$\pm$0.00} & 0.83 & 1235 & 0.06 & 534 \\
VMAS & 10 & 8.40{\tiny$\pm$0.02} & 8.50 & 1190 & 0.08 & 536 & 3.41{\tiny$\pm$0.02} & 3.47 & 2935 & 0.07 & 535 \\
VMAS & 25 & 22.9{\tiny$\pm$0.08} & 23.1 & 1091 & 0.09 & 537 & 8.63{\tiny$\pm$0.05} & 8.71 & 2897 & 0.06 & 536 \\
VMAS & 50 & 54.6{\tiny$\pm$0.15} & 55.0 & 916 & 0.14 & 538 & 20.3{\tiny$\pm$0.17} & 20.4 & 2465 & 0.07 & 537 \\
VMAS & 100 & 147{\tiny$\pm$0.19} & 148 & 681 & 0.28 & 540 & 55.0{\tiny$\pm$0.55} & 55.5 & 1819 & 0.09 & 538 \\
VMAS & 500$^\dagger$ & 3377{\tiny$\pm$258} & 3688 & 149 & 5.67 & 551 & 1180{\tiny$\pm$24.4} & 1239 & 424 & 0.34 & 546 \\
VMAS & 1000$^\dagger$ & 12602{\tiny$\pm$911} & 12945 & 80 & 15.96 & 560 & 4205{\tiny$\pm$192} & 4288 & 238 & 0.22 & 558 \\
\midrule
JaxNav & 1 & 0.22{\tiny$\pm$0.00} & 0.23 & 4616 & 3.19 & 493 & 0.18{\tiny$\pm$0.00} & 0.20 & 5483 & 3.29 & 493 \\
JaxNav & 10 & 1.28{\tiny$\pm$0.02} & 1.31 & 7784 & 4.04 & 531 & 0.76{\tiny$\pm$0.01} & 0.78 & 13160 & 3.98 & 523 \\
JaxNav & 25 & 3.75{\tiny$\pm$0.03} & 3.78 & 6669 & 4.56 & 574 & 1.66{\tiny$\pm$0.04} & 1.74 & 15047 & 4.37 & 563 \\
JaxNav & 50 & 10.0{\tiny$\pm$0.02} & 10.1 & 4983 & 5.21 & 674 & 3.23{\tiny$\pm$0.06} & 3.31 & 15465 & 4.78 & 646 \\
JaxNav & 100 & 50.0{\tiny$\pm$0.09} & 51.1 & 1999 & 6.54 & 846 & 6.64{\tiny$\pm$0.16} & 6.71 & 15072 & 5.70 & 796 \\
JaxNav & 500$^\dagger$ & 1279{\tiny$\pm$5.63} & 1316 & 391 & 22.89 & 3102 & 83.2{\tiny$\pm$4.83} & 91.6 & 6028 & 17.64 & 1900 \\
JaxNav & 1000$^\dagger$ & 4769{\tiny$\pm$47.7} & 4817 & 210 & 48.78 & 7732 & 260{\tiny$\pm$12.8} & 266 & 3849 & 33.80 & 2970 \\
\midrule
PyRoboSim & 1 & 1.89{\tiny$\pm$0.00} & 1.91 & 529 & 0.31 & 121 & 0.02{\tiny$\pm$0.00} & 0.02 & 64003 & 0.32 & 121 \\
PyRoboSim & 10 & 24.8{\tiny$\pm$0.12} & 25.2 & 403 & 0.32 & 122 & 0.17{\tiny$\pm$0.00} & 0.27 & 57674 & 0.35 & 121 \\
PyRoboSim & 25 & 76.9{\tiny$\pm$0.39} & 77.7 & 325 & 0.34 & 124 & 0.45{\tiny$\pm$0.01} & 0.78 & 55551 & 0.40 & 121 \\
PyRoboSim & 50 & 232{\tiny$\pm$0.97} & 233 & 216 & 0.36 & 126 & 1.07{\tiny$\pm$0.02} & 1.43 & 46690 & 0.47 & 122 \\
PyRoboSim & 100 & 818{\tiny$\pm$10.1} & 822 & 122 & 0.39 & 131 & 3.03{\tiny$\pm$0.04} & 3.35 & 33060 & 0.64 & 123 \\
PyRoboSim & 500$^\dagger$ & 21767{\tiny$\pm$766} & 22676 & 23 & 1.16 & 163 & 82.5{\tiny$\pm$0.65} & 130 & 6063 & 2.08 & 127 \\
PyRoboSim & 1000$^\dagger$ & 100510{\tiny$\pm$439} & 104449 & 10 & 2.70 & 202 & 307{\tiny$\pm$3.08} & 473 & 3261 & 6.60 & 133 \\
\midrule
MVSim & 1 & \textbf{0.03}{\tiny$\pm$0.00} & \textbf{0.03} & \textbf{30606} & \textbf{0.00} & \textbf{76} & \textbf{0.01}{\tiny$\pm$0.00} & \textbf{0.01} & \textbf{76127} & \textbf{0.00} & \textbf{76} \\
MVSim & 10 & 0.50{\tiny$\pm$0.00} & 0.59 & 19970 & \textbf{0.00} & \textbf{77} & \textbf{0.16}{\tiny$\pm$0.00} & 0.27 & \textbf{63459} & \textbf{0.00} & \textbf{76} \\
MVSim & 25 & 1.73{\tiny$\pm$0.01} & 2.02 & 14473 & \textbf{0.00} & \textbf{77} & 0.55{\tiny$\pm$0.00} & 0.83 & 45259 & \textbf{0.00} & \textbf{77} \\
MVSim & 50 & 4.48{\tiny$\pm$0.05} & 5.22 & 11169 & \textbf{0.00} & \textbf{79} & 1.58{\tiny$\pm$0.05} & 2.40 & 31591 & \textbf{0.00} & \textbf{78} \\
MVSim & 100 & 16.6{\tiny$\pm$0.13} & 21.2 & 6015 & \textbf{0.01} & \textbf{81} & 7.89{\tiny$\pm$0.18} & 12.9 & 12673 & \textbf{0.00} & \textbf{80} \\
MVSim & 500$^\dagger$ & 88.7{\tiny$\pm$0.17} & 115 & 5634 & \textbf{0.02} & \textbf{95} & 29.0{\tiny$\pm$0.08} & 52.1 & 17252 & \textbf{0.02} & \textbf{92} \\
MVSim & 1000$^\dagger$ & 280{\tiny$\pm$1.28} & 373 & 3575 & \textbf{0.05} & \textbf{114} & 116{\tiny$\pm$0.15} & 187 & 8630 & \textbf{0.04} & \textbf{108} \\
\bottomrule
\end{tabular}
\end{table}

\subsection{GPU Execution and Batched Simulation}
\label{app:efficiency_gpu}
This section evaluates the same scenario on an NVIDIA RTX~4090 GPU.
IR-SIM provides an optional batched backend in PyTorch that steps $B$ worlds of $N$ robots as device tensors, with the same kinematics, collision rule, and LiDAR ray casting as the CPU path.
The tables report its float32 mode, which uses the same precision as VMAS and JaxNav.

\paragraph{Single world.}
For one world, GPU execution is bound by kernel launches and data transfer rather than computation (Table~\ref{tab:efficiency_gpu}a).
With LiDAR, IR-SIM's backend steps one world in $1.9$ to $3.0\,\mathrm{ms}$ for every $N$ from $10$ to $1000$, the lowest GPU latency at every team size, whereas VMAS is $2$ to $4\times$ slower on the GPU than on a CPU core.

\begin{table}[t]
\caption{GPU execution on an RTX 4090 (IR-SIM's GPU backend in float32). (a) Step latency (ms) with one world per process; the CPU rows repeat Table~\ref{tab:appendix_efficiency_full}. (b) Batched throughput with LiDAR ($10^3$ agent steps per second) for $B$ worlds per call; the IR-SIM CPU row runs one world per core. Bold: best per column; $\uparrow$/$\downarrow$: higher/lower is better; OOM: out of memory.}
\label{tab:efficiency_gpu}\label{tab:efficiency_batch}
\centering
\scriptsize
\setlength{\tabcolsep}{4pt}
{\small\textit{(a) Single-world step latency (ms, $\downarrow$)}}\par\smallskip
\begin{tabular}{@{}lrrrr@{}}
\toprule
Configuration & $N=10$ & $N=100$ & $N=500$ & $N=1000$ \\
\midrule
\multicolumn{5}{@{}l}{\textit{with 64-beam LiDAR}} \\
IR-SIM, CPU & \textbf{0.90} & 16.6 & 81.8 & 178 \\
VMAS, CPU & 8.40 & 147 & 3377 & 12602 \\
JaxNav, CPU & 1.28 & 50.0 & 1279 & 4769 \\
IR-SIM, GPU backend & 1.85 & \textbf{2.99} & \textbf{2.89} & \textbf{3.02} \\
IR-SIM, CPU step + CUDA tensors & 0.94 & 16.1 & 67.9 & 147 \\
VMAS, GPU & 22.0 & 414 & 7212 & 27253 \\
JaxNav, GPU & 5.16 & 49.9 & 236 & 476 \\
\midrule
\multicolumn{5}{@{}l}{\textit{without LiDAR}} \\
IR-SIM, CPU & \textbf{0.23} & \textbf{1.37} & 19.9 & 59.4 \\
VMAS, CPU & 3.41 & 55.0 & 1180 & 4205 \\
JaxNav, CPU & 0.76 & 6.64 & 83.2 & 260 \\
IR-SIM, GPU backend & 0.40 & 1.65 & \textbf{1.52} & \textbf{1.54} \\
IR-SIM, CPU step + CUDA tensors & 0.45 & 2.23 & 16.7 & 49.5 \\
VMAS, GPU & 11.4 & 204 & 4113 & 15145 \\
JaxNav, GPU & 5.25 & 47.5 & 239 & 474 \\
\bottomrule
\end{tabular}
\par\smallskip
{\small\textit{(b) Batched throughput ($10^3$ agent steps/s, $\uparrow$)}}\par\smallskip
\begin{tabular}{@{}llrrrrr@{}}
\toprule
$N$ & Configuration & $B=1$ & $B=16$ & $B=128$ & $B=512$ & $B=1024$ \\
\midrule
\multirow{4}{*}{10} & IR-SIM, CPU (one world per core) & \textbf{11.3} & \textbf{127} & -- & -- & -- \\
 & IR-SIM, GPU backend & 5.4 & 77.9 & \textbf{551} & \textbf{1,762} & \textbf{2,051} \\
 & VMAS, GPU & 0.5 & 7.3 & 58.3 & 234 & 426 \\
 & JaxNav, GPU & 3.2 & 54.3 & 418 & 1,130 & 1,837 \\
\midrule
\multirow{4}{*}{100} & IR-SIM, CPU (one world per core) & 6.0 & 67.7 & -- & -- & -- \\
 & IR-SIM, GPU backend & \textbf{33.7} & \textbf{507} & \textbf{777} & \textbf{928} & \textbf{912} \\
 & VMAS, GPU & 0.3 & 4.2 & 32.9 & 111 & 165 \\
 & JaxNav, GPU & 3.6 & 55.8 & 364 & 673 & OOM \\
\bottomrule
\end{tabular}
\end{table}

\paragraph{Batched throughput.}
Table~\ref{tab:efficiency_batch}b reports throughput when stepping $B$ worlds per call and keeping the observations on the device, as in a GPU training loop.
With $1024$ worlds, IR-SIM reaches $2.05$ and $0.91$ million agent steps per second at $N=10$ and $N=100$, above JaxNav ($1.84$ million at $N=10$; $0.67$ million at $N=100$ with $512$ worlds, the largest batch that fits in memory) and VMAS ($0.43$ and $0.16$ million).

\begin{table}[t]
\caption{(a) Peak memory (MB): resident memory of one world on one CPU core, and device memory for $B=1024$ worlds per call (bold: best per column; $\uparrow$/$\downarrow$: higher/lower is better; OOM: out of memory). (b) Largest difference in robot state $\mathbf{x}$, LiDAR range $r$, and stop decisions between IR-SIM's GPU backend and its CPU path over 200 steps with a 64-beam LiDAR.}
\label{tab:efficiency_memory}\label{tab:efficiency_exactness}
\centering
\scriptsize
\setlength{\tabcolsep}{3pt}
\begin{minipage}[t]{0.5\linewidth}\centering
{\small\textit{(a) Peak memory (MB, $\downarrow$)}}\par\smallskip
\begin{tabular}{@{}lrrrr@{}}
\toprule
\multirow{2}{*}{Simulator} & \multicolumn{2}{c}{with LiDAR} & \multicolumn{2}{c}{without LiDAR} \\
\cmidrule(lr){2-3}\cmidrule(l){4-5}
 & $N=100$ & $N=1000$ & $N=100$ & $N=1000$ \\
\midrule
\multicolumn{5}{@{}l}{\textit{CPU, one world on one core}} \\
IR-SIM & 123 & 353 & 101 & 129 \\
MVSim & \textbf{81} & \textbf{114} & \textbf{80} & \textbf{108} \\
VMAS & 540 & 560 & 538 & 558 \\
JaxNav & 846 & 7,732 & 796 & 2,970 \\
PyRoboSim & 131 & 202 & 123 & 133 \\
\midrule
\multicolumn{5}{@{}l}{\textit{GPU, $B=1024$ worlds, with LiDAR}} \\
 & $N=10$ & $N=100$ & & \\
IR-SIM, float32 & 597 & 6,124 & & \\
IR-SIM, float64 & 1,153 & 11,918 & & \\
VMAS & \textbf{144} & \textbf{555} & & \\
JaxNav & 917 & OOM & & \\
\bottomrule
\end{tabular}
\end{minipage}\hfill
\begin{minipage}[t]{0.48\linewidth}\centering
\setlength{\tabcolsep}{2pt}
{\small\textit{(b) GPU backend vs.\ CPU path ($\downarrow$)}}\par\smallskip
\begin{tabular}{@{}llrrr@{}}
\toprule
Precision & $N$ & max $|\Delta\mathbf{x}|$ & max $|\Delta r|$ (m) & stop mismatches \\
\midrule
float64 & 10 & $3.6\times10^{-15}$ & $\mathbf{1.4\times10^{-13}}$ & 0 of 2,000 \\
float64 & 100 & $\mathbf{1.8\times10^{-15}}$ & $2.8\times10^{-13}$ & 0 of 20,000 \\
float32 & 10 & $9.6\times10^{-6}$ & $2.7\times10^{-4}$ & 0 of 2,000 \\
float32 & 100 & $1.6\times10^{-5}$ & $4.5$ & 0 of 20,000 \\
\bottomrule
\end{tabular}
\end{minipage}
\end{table}

\paragraph{Memory.}
On the CPU, IR-SIM uses $101$ to $353\,\mathrm{MB}$, compared with $80$ to $114\,\mathrm{MB}$ for MVSim, $538$ to $560\,\mathrm{MB}$ for VMAS and $796$ to $7{,}732\,\mathrm{MB}$ for JaxNav (Table~\ref{tab:efficiency_memory}a).
On the GPU with $1024$ worlds of $100$ robots, IR-SIM's backend uses $6{,}124\,\mathrm{MB}$ and VMAS $555\,\mathrm{MB}$, while JaxNav runs out of memory.

\paragraph{Exactness.}
In float64, the batched backend reproduces the CPU path: robot states agree to $4\times10^{-15}$, LiDAR ranges to $3\times10^{-13}\,\mathrm{m}$, and all collision decisions are identical (Table~\ref{tab:efficiency_exactness}b).

In float32, states agree to $2\times10^{-5}$ and maximum range differences are $2.7\times10^{-4}\,\mathrm{m}$ at $N=10$ and $4.54\,\mathrm{m}$ at $N=100$, with hit/miss flips on grazing beams.

\section{Scenario Construction Study}
\label{app:construction_full}

This section summarizes the briefs, the validator, the agent setup, and the complete results of the scenario-construction study in Section~\ref{sec:construction}.

\subsection{Briefs}
\label{app:construction_briefs}
Each brief is a machine-readable specification from which the text shown to the agent is rendered, so the text and the validator cannot disagree.
It fixes the world, the time step ($0.1\,\mathrm{s}$), the run length, the obstacles, the robots (differential drive, footprint, speed limit, sensors, and start pose), the task, and the pedestrians where present, and it ends with the list of acceptance checks.
Table~\ref{tab:appendix_construction_briefs} summarizes the thirteen reported briefs; family E contains rich scenes, with crowds, seeded random layouts, and limited field-of-view sensors, that need no task-specific control code in any simulator.

\begin{table}[t]
\caption{Scenario-construction briefs reported in Table~\ref{tab:construction_oneshot_opus55}. All briefs require constructing a scene from scratch. All robots use differential-drive kinematics and circular footprints.}
\label{tab:appendix_construction_briefs}
\centering
\scriptsize
\setlength{\tabcolsep}{3pt}
\begin{tabular}{@{}ll>{\raggedright\arraybackslash}p{0.68\linewidth}c@{}}
\toprule
Brief & Family & Scenario and task & Steps \\
\midrule
A1 & static & $12\times12$ m world, five rectangular obstacles; one robot ($r=0.25$ m, 90-beam $360^\circ$ LiDAR, 5 m) drives from $(1,1)$ to $(11,11)$ & 300 \\
A2 & static & $16\times10$ m world, three wall segments forming a zigzag corridor; one robot ($r=0.3$ m, 180-beam $270^\circ$ LiDAR, 8 m) passes five ordered waypoints and reaches its goal & 600 \\
B1 & pedestrians & $10\times10$ m world; four pedestrians ($r=0.2$ m, $0.6$ m/s) walk back and forth across the robot's path; the robot ($r=0.25$ m, no sensors) crosses from $(1,5)$ to $(9,5)$ without collision & 400 \\
B2 & pedestrians & $12\times12$ m world with a central $2\times2$ m pillar; eight pedestrians wander between random free-space targets at $0.5$ m/s; the robot ($r=0.3$ m) crosses the diagonal; the run must be reproducible under \texttt{--seed} & 500 \\
C1 & multi-robot & six robots ($r=0.2$ m, 36-beam LiDAR, 3 m) evenly placed on a circle of radius 4 m swap to diametrically opposite goals without colliding & 500 \\
C2 & multi-robot & $10\times10$ m warehouse with three shelves; three robots ($r=0.2$ m, $0.8$ m/s) cycle through their own four-waypoint loops, each waypoint at least twice, without colliding & 800 \\
E1 & scene & $16\times16$ m plaza with three boxes and three pillars; three robots ($r=0.25$ m, 90-beam LiDAR, 6 m) drive to goals across the plaza among twelve pedestrians wandering through the free space and four patrolling ones; no collision; seeded reproducibility & 600 \\
E2 & scene & $14\times12$ m warehouse with five shelves; three robots ($r=0.2$ m, 72-beam $270^\circ$ LiDAR) cycle through loops around shelves at least twice, two robots (36-beam $180^\circ$ LiDAR) patrol the open strips at least twice, two pedestrians wander inside the free strip on the right; seeded & 800 \\
E3 & scene & $14\times14$ m world with twelve circular pillars ($r=0.5$ m) placed uniformly at random inside the central $10\times10$ m square with $1.8$ m spacing using a seed; two robots ($r=0.3$ m, 60-beam LiDAR) cross diagonally without collision; identical layout for equal seeds & 500 \\
H1 & mixed & $16\times10$ m parking lot with seven parked cars; two robots ($r=0.25$ m, 48-beam $180^\circ$ LiDAR) drive to parking bays past three patrolling pedestrians & 500 \\
H2 & mixed & $15\times12$ m library with three shelves; one robot ($r=0.25$ m, 90-beam $270^\circ$ LiDAR) passes five waypoints through the aisles; three pedestrians wander in a reading area; seeded & 800 \\
H3 & mixed & $12\times8$ m, two rooms joined by a doorway; two robots ($r=0.25$ m, 60-beam LiDAR) cross via two waypoints each; two pedestrians patrol the rooms & 600 \\
H4 & mixed & $12\times10$ m, six robots ($r=0.2$ m, 24-beam LiDAR) swap between two lines past two pillars & 600 \\
\bottomrule
\end{tabular}
\end{table}

\subsection{Deliverable Contract and Validator}
\label{app:construction_validator}
The deliverable is the scenario in the simulator's native form (a YAML file for IR-SIM, a \texttt{BaseScenario} subclass for VMAS, and a \texttt{create\_world()} function for PyRoboSim) together with a script \texttt{run.py} that runs it headlessly and writes a rollout trace in a common JSON format.
The validator executes \texttt{run.py}, loads the native artifact through the simulator's API, and evaluates the brief's acceptance checks: simulator usage, world size and step count, entity counts, obstacle geometry, robot kinematics, footprint, start pose and LiDAR parameters, task completion, speed limits and world bounds, pedestrian motion, collision freedom, and, where required, determinism under a seed.
A deliverable is \emph{executable} if \texttt{run.py} succeeds and the trace has the requested shape, and \emph{valid} if every check passes; the \emph{specification accuracy} is the fraction of checks passed.

\subsection{Agent Configuration and Budget}
\label{app:construction_agent}
The agent is Claude Code in non-interactive mode with \texttt{claude-opus-5-5} or \texttt{claude-sonnet-5}, using file and search tools but no shell, web access, or sub-agents.
A read-only copy of the simulator's documentation and source is provided, and the skill condition additionally installs the five \texttt{ir-sim-skill} skills (version 0.9.1).
Each attempt writes the deliverable once within $60$ tool turns and is graded without repair.
Work directories are isolated, and the session transcripts confirm that no attempt accessed files outside them.

\subsection{Reference Solutions}
\label{app:construction_refs}
All thirteen briefs were solved in IR-SIM, VMAS, and PyRoboSim, and six compatible briefs were solved in JaxNav before the study to confirm that each is achievable and to test the validator.

\subsection{Coverage of Other Declarative Simulators}
\label{app:construction_coverage}
IR-SIM, VMAS, and PyRoboSim were chosen because all thirteen briefs lie within their common capabilities (Table~\ref{tab:appendix_construction_coverage}).

JaxNav~\citep{rutherford2024noregrets} is evaluated on the six briefs it can express (A1, A2, C1, C2, E3, H4), since all of its agents share one model and it has no moving obstacles.
GPUDrive~\citep{kazemkhani2025gpudrive} supports only vehicle models and fixed $91$-step episodes, and MVSim~\citep{blanco2023mvsim} and Flatland~\citep{flatland2026} require system or ROS installations, so they are not included.

\begin{table}[t]
\caption{Declarative simulators considered for the construction study and the requirements of the briefs they can express natively. Yes/no entries for the measured simulators were established by the reference solutions; the GPUDrive entries were read from its source (commit \texttt{aa48a43}); MVSim and Flatland entries come from their documentation and were not executed here.}
\label{tab:appendix_construction_coverage}
\centering
\scriptsize
\setlength{\tabcolsep}{3pt}
\newcolumntype{Y}{>{\raggedright\arraybackslash}p{0.105\linewidth}}
\begin{tabular}{@{}>{\raggedright\arraybackslash}p{0.095\linewidth}>{\raggedright\arraybackslash}p{0.135\linewidth}YYYY>{\raggedright\arraybackslash}p{0.08\linewidth}>{\raggedright\arraybackslash}p{0.09\linewidth}@{}}
\toprule
Simulator & Scene description & Diff.-drive, per-robot radius & Boxes and pillars anywhere & LiDAR beams / FOV & Scripted pedestrians (own radius) & 30--100 s runs at 0.1 s & Measured \\
\midrule
IR-SIM & YAML objects, behaviors, sensors & yes & yes & yes & yes & yes & 13 briefs \\
VMAS & Python scenario class & yes & yes & yes & yes & yes & 13 briefs \\
PyRoboSim & Python / YAML world & yes & yes (locations) & yes & yes (extra robots) & yes & 13 briefs \\
JaxNav & occupancy grid and agent states & one shared radius & yes (grid cells) & one shared model, $\geq 2$ beams & no & yes & 6 briefs \\
GPUDrive & Waymo-format JSON logs & vehicle models only & road polylines & fixed sample count & logged trajectories; non-vehicles removed by default & no (91 steps) & no \\
MVSim & XML world & yes & yes & yes & scripted vehicles & yes & not installable here \\
Flatland & YAML world and models & yes (plugins) & yes & yes (plugin) & scripted models & yes & ROS~1 only \\
\bottomrule
\end{tabular}
\end{table}

\subsection{Complete Results}
\label{app:construction_results}
Table~\ref{tab:construction_oneshot_opus55} aggregates $174$ attempts per model; none was stopped by an API rate limit.
With \texttt{claude-opus-5-5}, most failures are specification errors rather than simulator limits: plain IR-SIM fails mainly because the agent adds a \texttt{name} key to the YAML \texttt{world} block (four of five failures), the skill condition fails twice on E1 (a robot--pedestrian collision), and PyRoboSim fails three of four times by simulating the scene without importing \texttt{pyrobosim}.
With skills, attempts take an average of $98\,\mathrm{s}$ compared with $204$ to $238\,\mathrm{s}$ for the other conditions (Mann--Whitney test, $p<0.001$), while the validity gain over plain IR-SIM is not significant (Fisher's exact test, $p=0.43$).
The gap is largest on the rich scenes of family E ($135\,\mathrm{s}$ and \$$0.66$ against $290$ to $339\,\mathrm{s}$ and \$$1.35$ to \$$1.54$), where crowds, patrols, loops, and seeded layouts are YAML groups in IR-SIM but hand-written code elsewhere.
Because \texttt{ir-sim-skill} 0.9.1 was refined on failures from these brief families, the skill result is an upper bound rather than an independent measurement.

With \texttt{claude-sonnet-5}, the skills raise validity over plain IR-SIM from $61.5\%$ to $94.9\%$ (Fisher's exact test, $p<0.001$).

\subsection{Human Reference}
\label{app:construction_human}
Five participants authored brief A1 with the IR-SIM documentation through a single-page web study that follows the same single-shot protocol (Table~\ref{tab:construction_human}).
All five submissions are executable and three are valid. Authoring took $13$ to $25$ active minutes (mean $19.9$); the failures involve specification errors, such as a misplaced obstacle or an incorrect initial heading.
With five participants, these results indicate the scale of human authoring time rather than a statistical comparison.

\begin{table}[t]
\caption{Human single-shot submissions in IR-SIM, one row per participant, graded with the validator of the agent study: Docs: documentation only (docs) or documentation and the skill recipes (docs+skill); Fam.: self-rated IR-SIM familiarity (1--5) and years of Python experience; Time: active seconds from opening the brief to the submission (pauses excluded); Ran: the participant reported running the scene outside the page; Exec., Valid, Spec.\ acc.\ and the failed checks are defined as in the agent tables; Lines: non-comment Python / YAML lines. The row marked $^{\ddagger}$ in the single-shot table aggregates these.}
\label{tab:construction_human}
\centering
\scriptsize
\setlength{\tabcolsep}{2.5pt}
\begin{tabular}{llclrccccp{38mm}l}
\toprule
Participant & Docs & Fam.\ / Py & Task & Time (s)$\downarrow$ & Ran & Exec. & Valid & Spec.\ acc.\ (\%)$\uparrow$ & Failed checks & Lines \\
\midrule
P05 & docs & 3 / 7 & A1 & 768 & no & yes & yes & 100 & -- & 69 / 35 \\
P100 & docs & 4 / 6 & A1 & 1494 & yes & yes & no & 83 & \texttt{obstacles\_present}, \texttt{collision\_free} & 69 / 40 \\
P11 & docs & 4 / 6 & A1 & 1152 & yes & yes & yes & 100 & -- & 69 / 40 \\
p16 & docs & 3 / 8 & A1 & 1380 & no & yes & no & 75 & \texttt{robots\_start}, \texttt{robots\_lidar}, \texttt{collision\_free} & 69 / 25 \\
P24 & docs & 4 / 8 & A1 & 1176 & no & yes & yes & 100 & -- & 69 / 30 \\
\bottomrule
\end{tabular}
\end{table}

\section{Social Navigation Benchmark Protocol}
\label{app:social_benchmark}

This section lists the settings behind Table~\ref{tab:social_benchmark}; Table~\ref{tab:social_benchmark_full} gives every metric for every controller and team size.

\paragraph{Scenario.}
Robots have random convex polygonal footprints (average radius $0.1$--$0.4\,\mathrm{m}$), random starts and goals, no static obstacles, velocity bounds $|v|\le 2\,\mathrm{m/s}$ and $|\omega|\le 3.14\,\mathrm{rad/s}$, a $0.1\,\mathrm{s}$ step, a $500$-step horizon, and a $0.4\,\mathrm{m}$ arrival threshold; a collision stops the robots involved.
Every episode reseeds the simulator from the base seed and the episode index, so all controllers see the same layouts.
An episode with any collision counts as a collision, and one in which some robot has neither arrived nor collided at the horizon counts as a timeout.

\paragraph{Controllers.}
ORCA and AVOCADO use a $15\,\mathrm{m}$ neighbor distance, five neighbors, a $0.5\,\mathrm{s}$ time horizon, a $0.1\,\mathrm{m}$ safety buffer, and a $1.0\,\mathrm{m/s}$ preferred speed (AVOCADO with its default opinion-dynamics parameters); VO-polytope runs the released implementation in reciprocal mode with a $5\,\mathrm{m}$ neighbor region; RDA runs one model predictive controller per robot (horizon $5$, at most five obstacles, two ADMM iterations, $0.2\,\mathrm{m}$ safety distance) without coordination.
SARL, CADRL, and LSTM-RL are trained in the original CrowdSim environment with the repository's default recipe ($3000$ imitation learning and $10\,000$ reinforcement learning episodes) and evaluated with their original action selection; RL-RVO uses the released pretrained policy without retraining.
ORCA, AVOCADO, and RL-RVO represent every robot by the largest enclosing circle in the scene, the CrowdNav policies observe each robot's own enclosing circle, and VO-polytope and RDA use the polygons directly.
Planar velocity outputs $\mathbf{v}$ from controllers are mapped to a differential-drive command by $v=\|\mathbf{v}\|\max(0,\cos\Delta\theta)$ and $\omega=\mathrm{clip}(3\Delta\theta,-\omega_{\max},\omega_{\max})$, where $\Delta\theta$ is the heading error between the robot and $\mathbf{v}$.

\paragraph{Runtime.}
All rows use the IR-SIM build of Section~\ref{sec:efficiency} on one core of an AMD Ryzen~9~7950X3D.
A $100$-episode cell, including policy inference, takes seconds to minutes for the reactive planners and tens of minutes for the per-robot MPC and the CrowdNav value-network policies (from $8\,\mathrm{s}$ for ORCA with five robots to about $110\,\mathrm{min}$ for RDA with ten robots).

\begin{table}[t]
\caption{Complete social navigation benchmark results behind Table~\ref{tab:social_benchmark}: episode-level outcome rates (\%), and navigation time and path length of robots that reach their goals and speed over active robot frames (mean $\pm$ population standard deviation), for every controller and team size. Bold: best value per team size; $\uparrow$/$\downarrow$: higher/lower is better.}
\label{tab:social_benchmark_full}
\centering
\scriptsize
\setlength{\tabcolsep}{4pt}
\renewcommand{\arraystretch}{0.95}
\resizebox{\linewidth}{!}{
\begin{tabular}{cl|ccc|ccc}
\toprule
Robots & Method &
Success$\uparrow$ & Collision$\downarrow$ & Timeout$\downarrow$ &
Time (s)$\downarrow$ & Speed (m/s)$\uparrow$ & Path (m)$\downarrow$ \\
\midrule
5 & ORCA~\citep{van2011reciprocal} & \textbf{100.0} & \textbf{0.0} & \textbf{0.0} & $5.05\pm2.47$ & $\mathbf{0.897}\pm0.125$ & $4.60\pm2.26$ \\
5 & VO-polytope~\citep{huang2023velocity} & 83.0 & 17.0 & \textbf{0.0} & $\mathbf{4.74}\pm2.28$ & $0.888\pm0.135$ & $4.38\pm2.21$ \\
5 & AVOCADO~\citep{martinez2025avocado} & \textbf{100.0} & \textbf{0.0} & \textbf{0.0} & $5.23\pm2.38$ & $0.827\pm0.139$ & $4.49\pm2.21$ \\
5 & RDA~\citep{han2023rda} & 20.0 & \textbf{0.0} & 80.0 & $5.03\pm3.48$ & $0.601\pm0.400$ & $\mathbf{4.14}\pm2.14$ \\
5 & SARL~\citep{chen2019crowdrobot} & 41.0 & \textbf{0.0} & 59.0 & $10.90\pm7.80$ & $0.589\pm0.222$ & $6.62\pm4.14$ \\
5 & CADRL~\citep{chen2017decentralized} & 9.0 & 6.0 & 85.0 & $8.79\pm6.87$ & $0.578\pm0.219$ & $5.12\pm3.07$ \\
5 & LSTM-RL~\citep{everett2018motion} & 39.0 & 13.0 & 48.0 & $9.97\pm6.92$ & $0.632\pm0.199$ & $6.46\pm3.94$ \\
5 & RL-RVO~\citep{han2022reinforcement} & 89.0 & 6.0 & 5.0 & $5.82\pm3.46$ & $0.827\pm0.144$ & $4.84\pm2.47$ \\
\midrule
10 & ORCA~\citep{van2011reciprocal} & \textbf{90.0} & 1.0 & 9.0 & $5.41\pm3.04$ & $\mathbf{0.850}\pm0.151$ & $4.63\pm2.40$ \\
10 & VO-polytope~\citep{huang2023velocity} & 45.0 & 55.0 & \textbf{0.0} & $\mathbf{4.31}\pm2.44$ & $0.835\pm0.172$ & $3.91\pm2.33$ \\
10 & AVOCADO~\citep{martinez2025avocado} & 86.0 & \textbf{0.0} & 14.0 & $5.50\pm2.75$ & $0.783\pm0.160$ & $4.49\pm2.33$ \\
10 & RDA~\citep{han2023rda} & 0.0 & \textbf{0.0} & 100.0 & $5.94\pm6.04$ & $0.356\pm0.346$ & $\mathbf{3.64}\pm2.44$ \\
10 & SARL~\citep{chen2019crowdrobot} & 26.0 & 11.0 & 63.0 & $10.96\pm8.41$ & $0.543\pm0.215$ & $5.89\pm3.62$ \\
10 & CADRL~\citep{chen2017decentralized} & 0.0 & 40.0 & 60.0 & $8.93\pm7.30$ & $0.558\pm0.192$ & $4.90\pm3.20$ \\
10 & LSTM-RL~\citep{everett2018motion} & 0.0 & 36.0 & 64.0 & $12.69\pm11.02$ & $0.535\pm0.186$ & $7.01\pm5.38$ \\
10 & RL-RVO~\citep{han2022reinforcement} & 73.0 & 20.0 & 7.0 & $6.87\pm4.19$ & $0.750\pm0.144$ & $5.18\pm2.94$ \\
\bottomrule
\end{tabular}
}
\end{table}

\section{Policy Training Protocol}
\label{app:policy_training}
This section lists the settings behind Table~\ref{tab:policy_training_pipeline}.

\paragraph{Port.}
CrowdNav's learner (trainer, explorer, replay memory, value networks, and schedule) is used without modification; only its environment class is replaced by a subclass that keeps CrowdNav's episode sampler, seeding, observation, reward, collision test, and termination, and delegates the state update to an IR-SIM world generated from CrowdNav's \texttt{env.config} (a $0.25\,\mathrm{s}$ step, a holonomic robot of radius $0.3\,\mathrm{m}$, and $N$ ORCA-driven humans of radius $0.3\,\mathrm{m}$ with CrowdNav's ORCA settings).
Because the humans do not observe the robot, the crowd commands are evaluated once per simulation step and reused for the $81$ one-step lookaheads of CrowdNav's action selection, which is exact.
Replaying $500$ test episodes with recorded actions reproduces every human state, reward, and termination step of \texttt{CrowdSim} exactly with five humans; with ten humans the training runs of the two simulators differ slightly (Table~\ref{tab:policy_training_pipeline_full}).

\paragraph{Training.}
All runs use CrowdNav's default recipe: $3000$ imitation-learning episodes from ORCA, then $10\,000$ reinforcement-learning episodes with $100$ SGD batches after each episode, $\epsilon$-greedy exploration decaying from $0.5$ to $0.1$, target-network updates every $50$ episodes, validation every $1000$ episodes, and a final test on $500$ fixed cases.
Each (policy, humans, seed) configuration is trained once in \texttt{CrowdSim} and once in IR-SIM with the same seed, and therefore the same episode layouts and exploration draws; the two runs of a pair train concurrently on two dedicated cores each, all on the same CPU chiplet.

\paragraph{Batched TorchRL implementation (C, D).}
The environment is a TorchRL \texttt{EnvBase} with a batch of $16$ worlds, each using the IR-SIM port (D) or the unmodified \texttt{CrowdSim} (C) and starting the next training case as soon as its episode ends; the policy scores the $81$ candidate actions of all worlds in one pass of the SARL value network, which reproduces the action of CrowdNav's \texttt{predict} in every checked decision.
The recipe is CrowdNav's (same demonstrations, value targets, exploration schedule, SGD batches, target updates, validation, and test), so C and D differ from A and B only in the sampling order of the concurrent worlds, not in the algorithm; C evaluates the crowd twice per step where D evaluates it once, and D (GPU) runs the network passes and SGD updates on the RTX~4090.

\paragraph{Measurements.}
Simulator time per decision is the time spent in \texttt{step} and \texttt{onestep\_lookahead} over the reinforcement-learning phase divided by the number of decisions (for C and D, the batched environment step, its observation, and the vectorized one-step reward); network time is the remaining rollout time per decision, i.e., the value-network passes and action selection ($81$ sequential calls per decision in A and B, one batched call per step shared by all worlds in C and D); rollout time covers the training and validation episodes of that phase, and the total adds imitation learning, the SGD updates, and the final test.

\paragraph{Additional runs.}
Table~\ref{tab:policy_training_pipeline_full} repeats all configurations with a second seed and with ten humans; every trained policy achieves approximately $100\%$ success on the $500$ test episodes (rates rounded to whole percentages).

With the second seed, the batched implementation with IR-SIM (\texttt{CrowdSim}) is $8.4\times$ ($8.5\times$) faster than the original CrowdNav--CrowdSim pipeline, and $11.4\times$ with the network on the GPU, which confirms the result of the main text.
With ten humans, the simulator accounts for a larger share of the cost: replacing only the simulator shortens training from $13.11$ to $7.78\,\mathrm{h}$ ($1.7\times$), the batched implementation needs $0.72\,\mathrm{ms}$ of simulator time per decision with IR-SIM and $1.41\,\mathrm{ms}$ with \texttt{CrowdSim} ($1.27$ versus $1.44\,\mathrm{h}$ of training), and the GPU variant trains in $0.70\,\mathrm{h}$ ($18.8\times$).
With five humans, \texttt{CrowdSim}'s step is slightly cheaper than IR-SIM's ($0.36$ versus $0.43\,\mathrm{ms}$) and the two batched runs take the same time.

\begin{table}[t]
\centering
\small
\caption{Separating the simulator from the training implementation (SARL; five humans with seeds $0$ and $1$, ten humans with seed $0$; speed-up relative to A in the same group). A and B run CrowdNav's original training code with \texttt{CrowdSim} and IR-SIM; C and D run the batched TorchRL implementation that steps $16$ worlds together and scores the candidate actions of all worlds in one network pass. Each run uses two dedicated CPU cores on the same chiplet; D (GPU) additionally runs the network on an RTX~4090. Sim.\ and net.\ time are the simulator and value-network time of one robot decision as in Table~\ref{tab:policy_training_pipeline}. Bold: best value per group; $\uparrow$/$\downarrow$: higher/lower is better.}
\label{tab:policy_training_pipeline_full}
\resizebox{\linewidth}{!}{%
\begin{tabular}{rr llll rrrr rrr}
\toprule
Humans & Seed & Config & Method & Simulator & Hardware & Sim.\ time (ms)$\downarrow$ & Net.\ time (ms)$\downarrow$ & Time (h)$\downarrow$ & Speed-up$\uparrow$ & Success (\%)$\uparrow$ & Collision (\%)$\downarrow$ & Nav.\ time (s)$\downarrow$ \\
\midrule
5 & 0 & A & CrowdNav & \texttt{CrowdSim} & CPU & 6.86 & 27.77 & 6.99 & 1.0$\times$ & 100 & 0 & 10.54 \\
5 & 0 & B & CrowdNav & IR-SIM & CPU & 2.50 & 27.00 & 6.13 & 1.1$\times$ & 100 & 0 & 10.54 \\
5 & 0 & C & Batched \texttt{CrowdSim} + TorchRL & \texttt{CrowdSim} & CPU & \textbf{0.36} & 0.44 & 0.74 & 9.4$\times$ & 100 & 0 & \textbf{10.38} \\
5 & 0 & D & Batched IR-SIM + TorchRL & IR-SIM & CPU & 0.43 & 0.47 & 0.76 & 9.2$\times$ & 100 & 0 & \textbf{10.38} \\
5 & 0 & D (GPU) & Batched IR-SIM + TorchRL & IR-SIM & CPU + GPU & 0.41 & \textbf{0.11} & \textbf{0.51} & \textbf{13.7$\times$} & 100 & 0 & 10.57 \\
\midrule
5 & 1 & A & CrowdNav & \texttt{CrowdSim} & CPU & 6.98 & 28.10 & 7.07 & 1.0$\times$ & 100 & 0 & 10.64 \\
5 & 1 & B & CrowdNav & IR-SIM & CPU & 2.59 & 27.47 & 6.25 & 1.1$\times$ & 100 & 0 & 10.64 \\
5 & 1 & C & Batched \texttt{CrowdSim} + TorchRL & \texttt{CrowdSim} & CPU & \textbf{0.38} & 0.49 & 0.83 & 8.5$\times$ & 100 & 0 & \textbf{10.59} \\
5 & 1 & D & Batched IR-SIM + TorchRL & IR-SIM & CPU & 0.47 & 0.49 & 0.84 & 8.4$\times$ & 100 & 0 & \textbf{10.59} \\
5 & 1 & D (GPU) & Batched IR-SIM + TorchRL & IR-SIM & CPU + GPU & 0.45 & \textbf{0.12} & \textbf{0.62} & \textbf{11.4$\times$} & 100 & 0 & 10.65 \\
\midrule
10 & 0 & A & CrowdNav & \texttt{CrowdSim} & CPU & 31.17 & 29.43 & 13.11 & 1.0$\times$ & 100 & 0 & 11.91 \\
10 & 0 & B & CrowdNav & IR-SIM & CPU & 4.01 & 28.17 & 7.78 & 1.7$\times$ & 100 & 0 & \textbf{11.64} \\
10 & 0 & C & Batched \texttt{CrowdSim} + TorchRL & \texttt{CrowdSim} & CPU & 1.41 & 0.75 & 1.44 & 9.1$\times$ & 100 & 0 & 12.44 \\
10 & 0 & D & Batched IR-SIM + TorchRL & IR-SIM & CPU & 0.72 & 0.76 & 1.27 & 10.3$\times$ & 100 & 0 & 12.02 \\
10 & 0 & D (GPU) & Batched IR-SIM + TorchRL & IR-SIM & CPU + GPU & \textbf{0.70} & \textbf{0.12} & \textbf{0.70} & \textbf{18.8$\times$} & 100 & 0 & 12.25 \\
\bottomrule
\end{tabular}}
\end{table}

\section{Simulator Bridges and Robot Interfaces}
\label{sec:bridges-interfaces}

IR-SIM is designed for fast 2D scenario construction, but many navigation studies still require richer visual context, simulator-specific sensors, or physical robot validation.
To support this workflow, IR-SIM provides bridge interfaces that preserve lightweight scenario logic while delegating perception, rendering, and physics simulation to external platforms.

\begin{figure}[t]
    \centering
    \includegraphics[width=\linewidth]{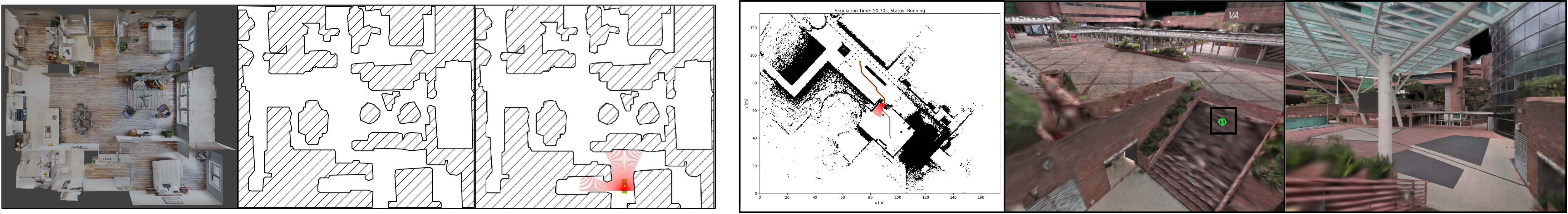}\\[1pt]
    {\scriptsize (a) Habitat-Sim/HM3D to IR-SIM occupancy map conversion (left) and IR-SIM trajectory and map in a Gaussian Splatting scene (right)}\\[5pt]
    \includegraphics[width=\linewidth]{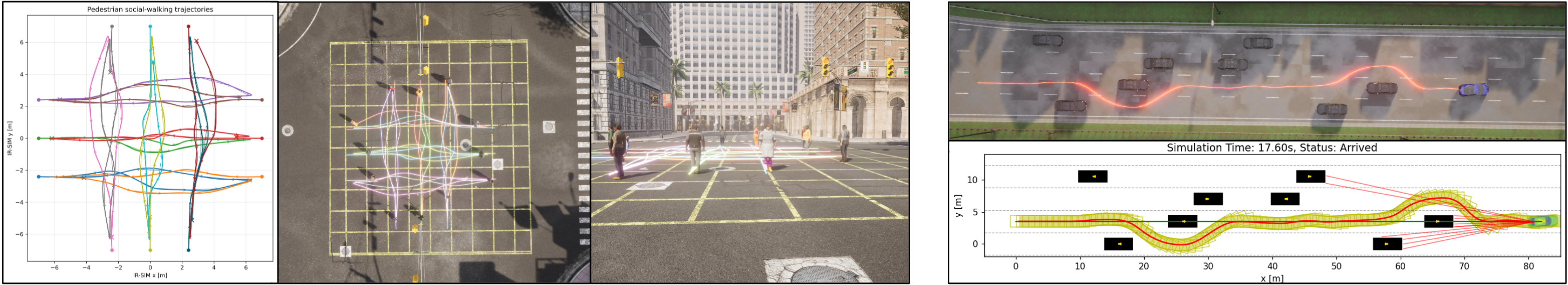}\\[1pt]
    {\scriptsize (b) IR-SIM pedestrian trajectories instantiated in CARLA (left) and NeuPAN control executed through IR-SIM with CARLA sensing and map information (right)}\\[5pt]
    \includegraphics[width=\linewidth]{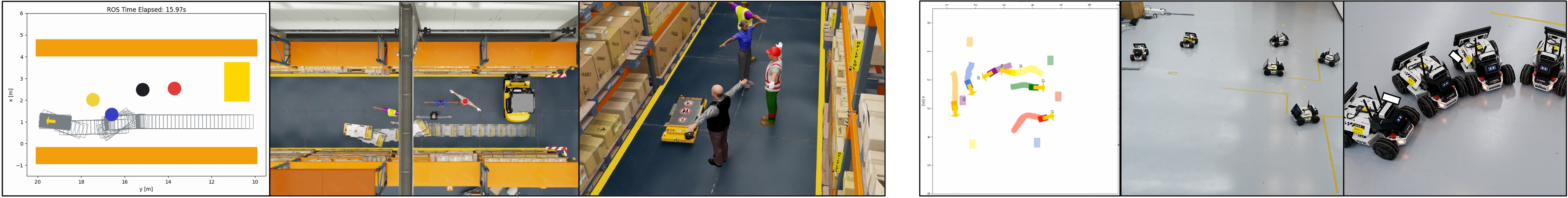}\\[1pt]
    {\scriptsize (c) IR-SIM pedestrian trajectories synchronized with Isaac Sim (left) and IR-SIM state and reference trajectory publishing for a real robot (right)}
    \caption{Bridge examples: (a) 3D scene assets, (b) CARLA, and (c) deployment to Isaac Sim and physical robots.}
    \label{fig:bridges}
    \figpanellabel{fig:bridge_habitat_gs}
    \figpanellabel{fig:bridge_carla}
    \figpanellabel{fig:bridge_isaac_real}
\end{figure}

\subsection{Habitat-Sim Map Conversion}

The left side of Fig.~\ref{fig:bridge_habitat_gs} illustrates the conversion from Habitat-Sim/HM3D~\citep{savva_habitat_2019,ramakrishnan2021hm3d} to IR-SIM, where a 3D indoor scene is converted into a 2D IR-SIM occupancy map.
This conversion allows rich indoor assets to be reused for 2D path planning, sensor simulation, and navigation benchmarking without requiring a full 3D simulation loop.

\subsection{Gaussian Splatting Visualization}

The right side of Fig.~\ref{fig:bridge_habitat_gs} shows the transfer from IR-SIM to Gaussian Splatting~\citep{kerbl20233d}, where IR-SIM provides the navigation map and trajectory and the reconstructed 3D scene provides visual context for inspection.

\subsection{CARLA Integration}

The remaining bridge examples transfer IR-SIM scenarios or control policies into high-fidelity simulators and physical robots.
As shown on the left side of Fig.~\ref{fig:bridge_carla}, pedestrian trajectories generated by IR-SIM are instantiated in a CARLA street scene to add crowd interaction.
The right side of Fig.~\ref{fig:bridge_carla} shows NeuPAN control transferred from IR-SIM to CARLA, where CARLA provides LiDAR and map information while the NeuPAN policy~\citep{han2025neupan} is executed through IR-SIM.

\subsection{Isaac Sim Integration}

The left side of Fig.~\ref{fig:bridge_isaac_real} shows an Isaac Sim warehouse scene, where pedestrian trajectories generated by IR-SIM using RVO behavior~\citep{vandenberg2008reciprocal} are synchronized with the 3D simulator via ROS 2.

\subsection{Physical Robot Interfaces}

Finally, the right side of Fig.~\ref{fig:bridge_isaac_real} applies the same idea to physical robots by publishing the robot state and reference trajectory through ROS.
Together, these examples show how IR-SIM remains the lightweight scenario and control layer while external platforms provide richer rendering, sensing, or deployment environments, allowing a scenario authored once in IR-SIM to be reused across validation environments without reimplementing the scenario logic.

\section{Scenario Examples}

\begin{figure}[!htb]
    \centering
    \setlength{\tabcolsep}{3pt}
    \begin{tabular}{cc}
        \includegraphics[width=0.47\linewidth]{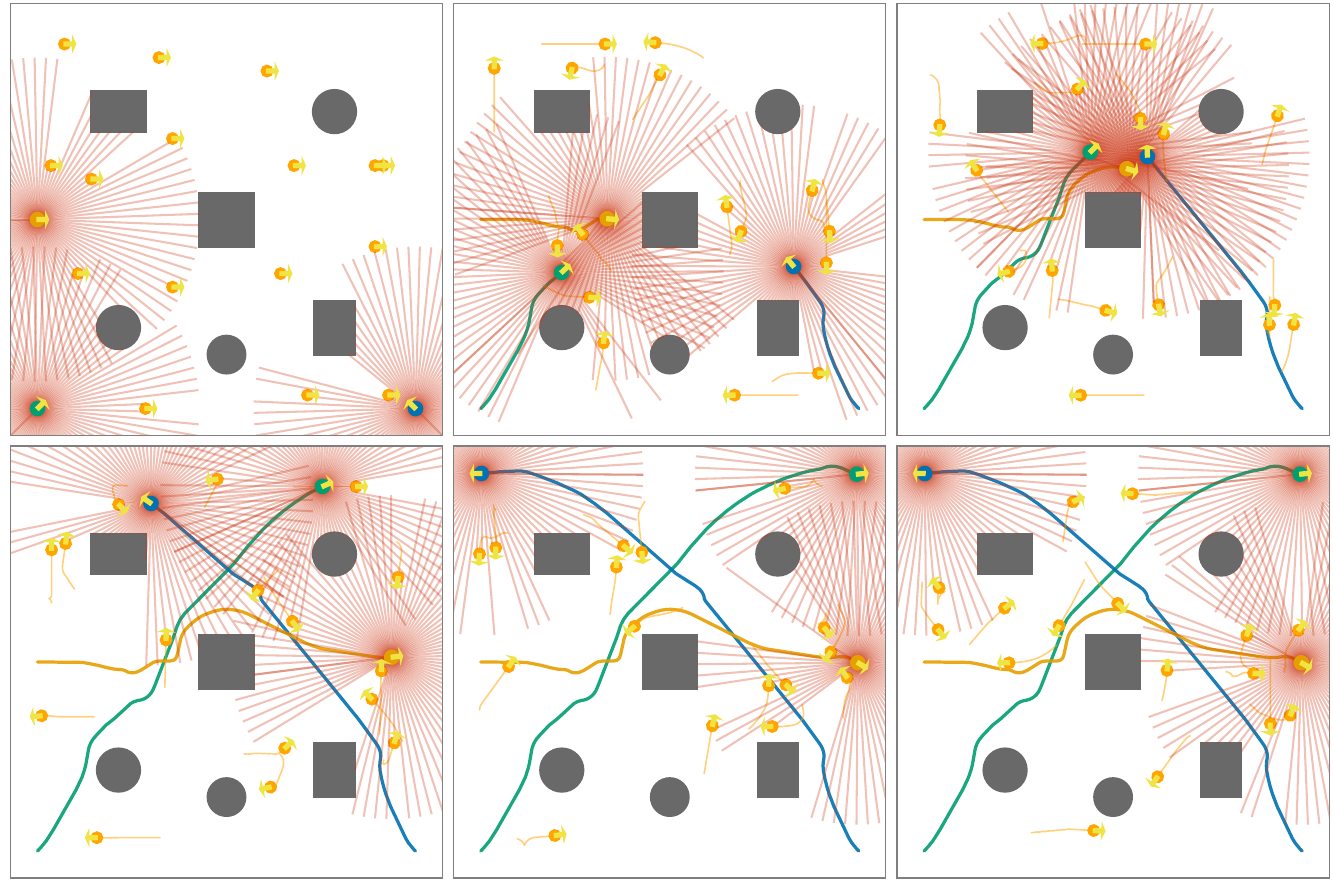} &
        \includegraphics[width=0.47\linewidth]{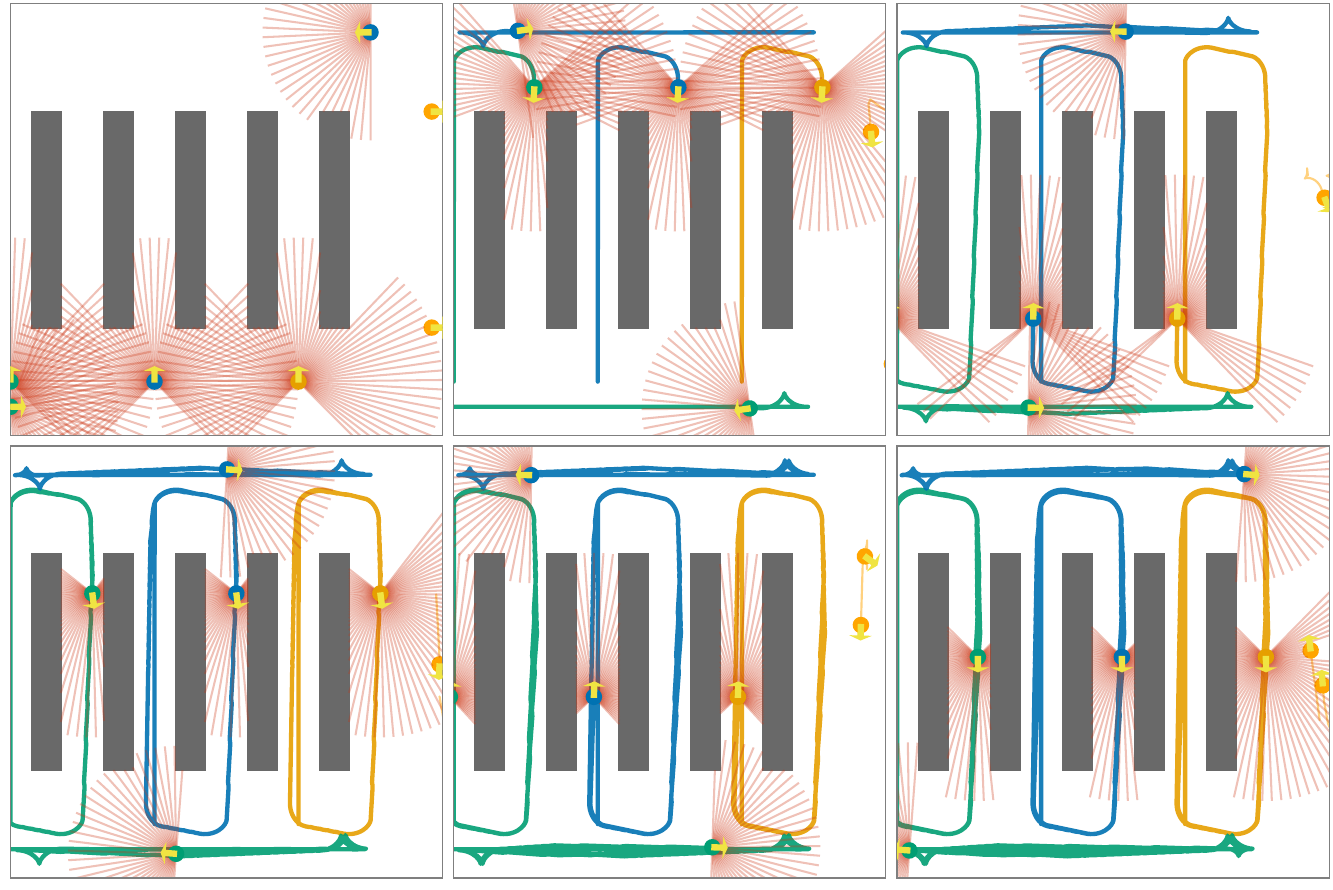} \\
        {\scriptsize (a) Plaza crowd: three robots among sixteen pedestrians} & {\scriptsize (b) Warehouse fleet with $270^\circ$ and $180^\circ$ LiDARs} \\[3pt]
        \includegraphics[width=0.47\linewidth]{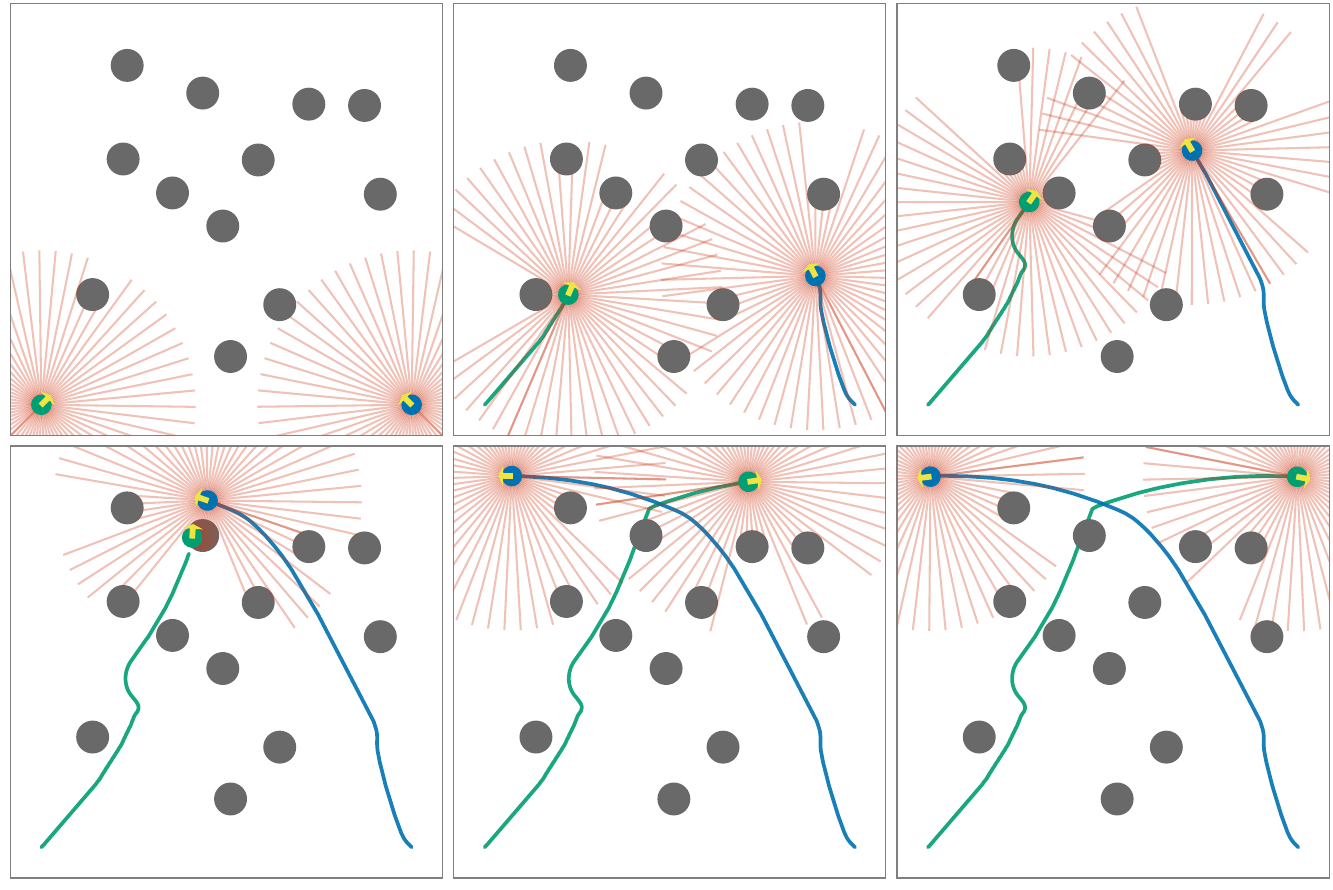} &
        \includegraphics[width=0.47\linewidth]{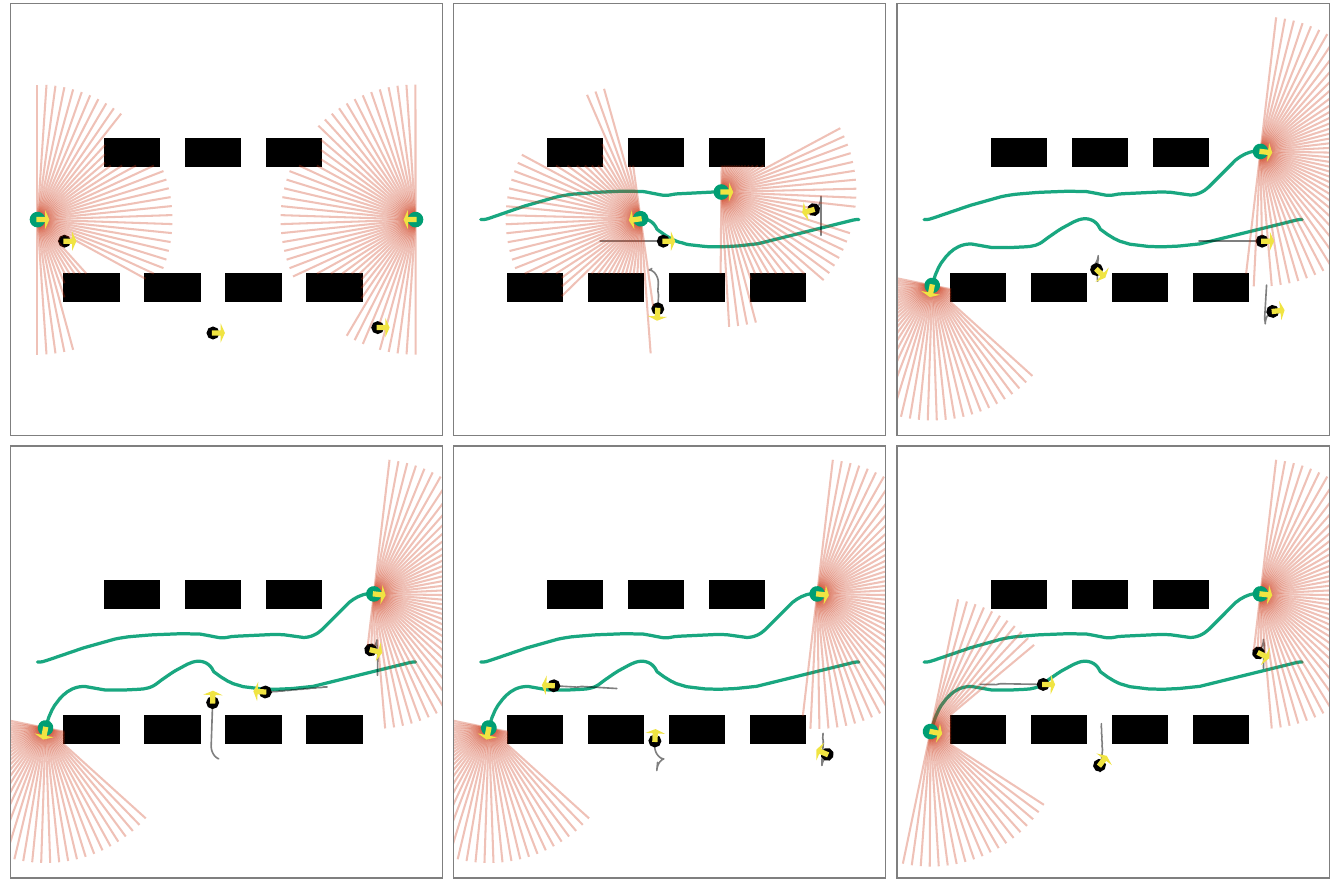} \\
        {\scriptsize (c) Seeded random pillar field} & {\scriptsize (d) Parking lot with patrolling pedestrians} \\[3pt]
        \includegraphics[width=0.47\linewidth]{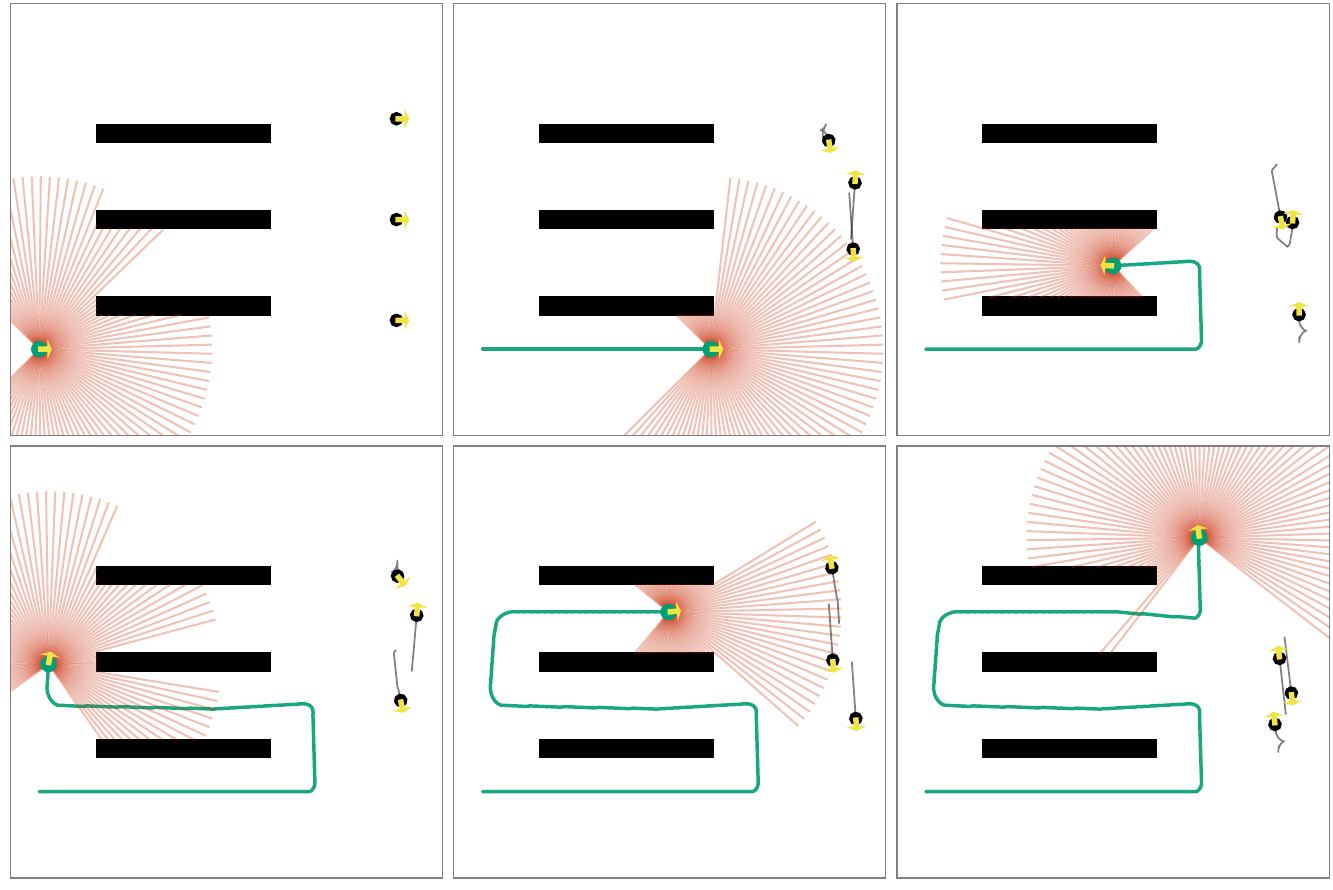} &
        \includegraphics[width=0.47\linewidth]{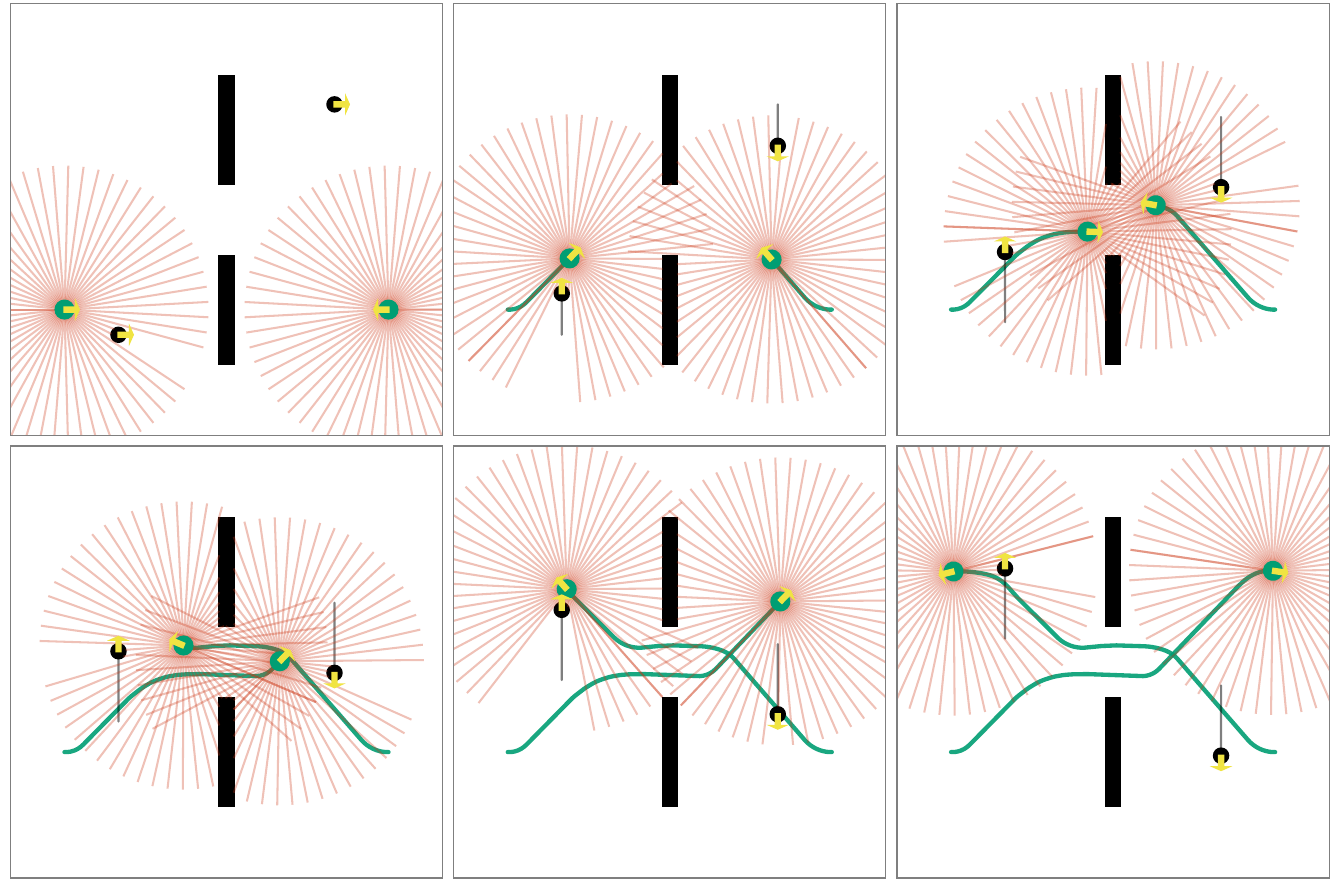} \\
        {\scriptsize (e) Library aisles with wandering readers} & {\scriptsize (f) Two rooms joined by a doorway} \\[3pt]
        \includegraphics[width=0.47\linewidth]{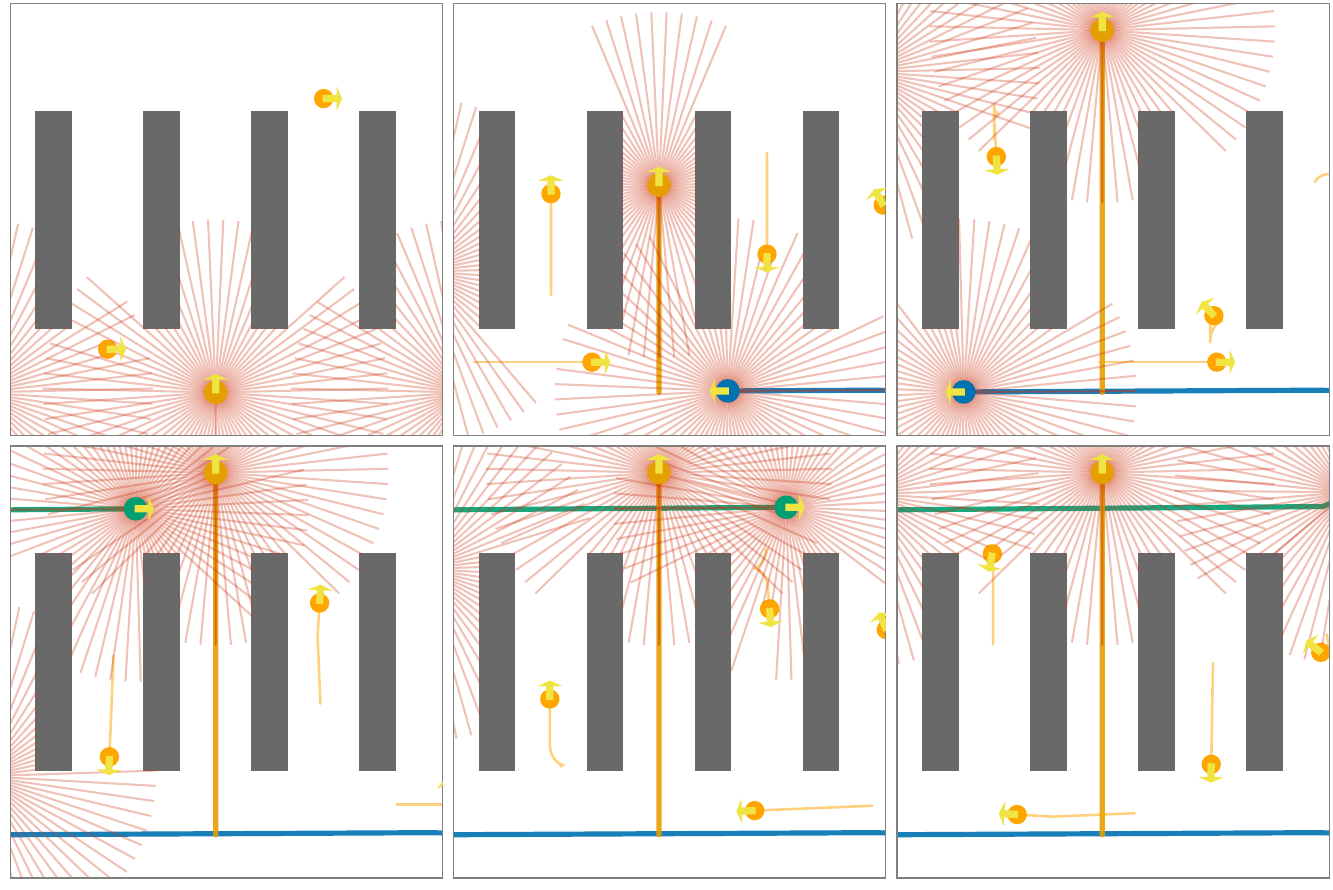} &
        \includegraphics[width=0.47\linewidth]{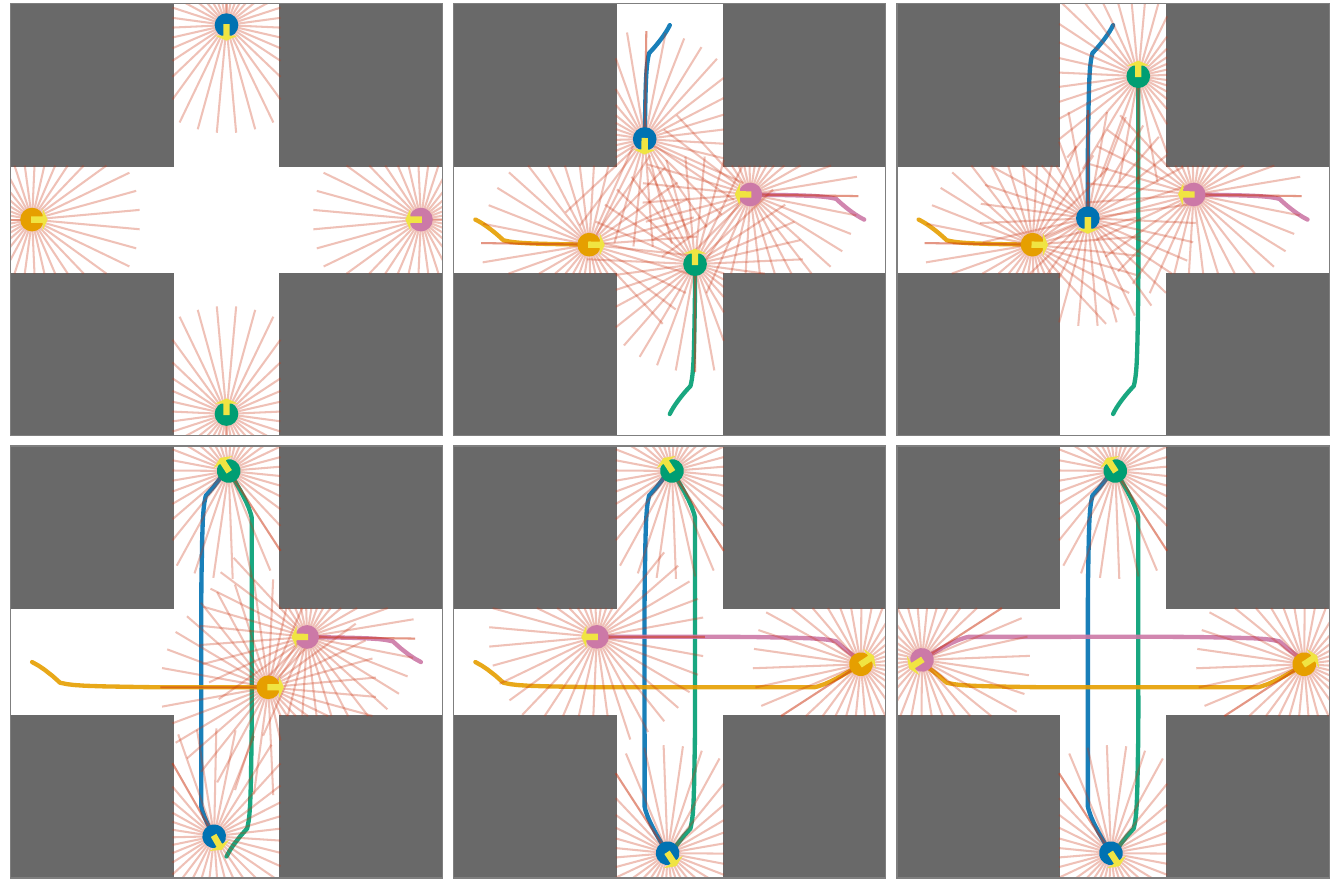} \\
        {\scriptsize (g) Warehouse crossing among patrols and wanderers} & {\scriptsize (h) Intersection square admitting one robot at a time}
    \end{tabular}
    \caption{Rollout snapshots (six per scene) of complex IR-SIM scenes replayed from recorded rollouts: robots with their LiDAR beams and trajectories, pedestrians in orange with short trails, static obstacles in gray.}
    \label{fig:app_complex_scenes}
    \figpanellabel{fig:app_complex_E1}
    \figpanellabel{fig:app_complex_E2}
    \figpanellabel{fig:app_complex_E3}
    \figpanellabel{fig:app_complex_H1}
    \figpanellabel{fig:app_complex_H2}
    \figpanellabel{fig:app_complex_H3}
    \figpanellabel{fig:app_complex_D1}
    \figpanellabel{fig:app_complex_D2}
\end{figure}

Fig.~\ref{fig:app_complex_scenes} shows six rollout snapshots of each of eight IR-SIM scenes described in YAML, the reference scenes of the construction study (Appendix~\ref{app:construction_full}) replayed from their recorded rollouts: crowded plazas, warehouse fleets with field-of-view LiDARs, seeded random layouts, parking lots, libraries, rooms joined by doorways, and intersections with patrolling and wandering pedestrians.

\end{document}